\documentclass[runningheads]{llncs}
\usepackage{booktabs}
\usepackage{amsmath}
\usepackage{amssymb}
\usepackage{tikz}
\usepackage{graphicx}
\usepackage{tabularx}
\usepackage{booktabs}
\usepackage{algorithm}
\usepackage{algpseudocode}
\usepackage{xcolor}

\usetikzlibrary{positioning,shapes.geometric,arrows.meta,fit,backgrounds,calc}
\graphicspath{{figures/}}

\usepackage{bm}

\usepackage{listings}
\lstdefinestyle{decisionlist}{
  language=Python,
  basicstyle=\ttfamily\scriptsize,
  breaklines=true,
  columns=fullflexible,
  frame=single,
}

\begin{document}

\title{Making Strides in Symbolic Imitation Learning: Offline Demonstrations, Online Tuning}

\titlerunning{Making Strides in Symbolic Imitation Learning}

\author{Bryce MacInnis\inst{1}\orcidID{0009-0001-6462-2058} \and Malcolm I. Heywood\inst{1}\orcidID{0000-0002-1521-0671}}

\authorrunning{MacInnis and Heywood}

\institute{Faculty of Computer Science, Dalhousie University, Halifax NS, Canada\\
\email{\{bmacinnis,mheywood\}@dal.ca}}

\maketitle

\begin{abstract}
Deep reinforcement learning (DRL) methods have been the dominant paradigm in continuous control tasks due to their efficacy in function approximation, while interpretable evolutionary methods have lagged largely behind for the opposite reason. In this work, we introduce Symbolic Imitation Learning, a framework for distilling an oracle policy through an offline dataset into a decision list with linear model motion primitives. We show that this, in conjunction, with online Augmented-Random Search (ARS) tuning of the constants and motion primitives approximates the performance of the oracle across 20 benchmark RL tasks and exceeds the oracle in 11 of those 20 tasks while simultaneously producing parsimonious, interpretable policies that domain experts can better understand.

\end{abstract}

\section{Introduction}

Reinforcement learning (RL) tasks require an agent to interact with an environment. As such the agent has to both learn what is worth learning from and learn an optimal policy for solving the task \cite{sutton18}. The goal of RL is generally expressed as reward maximization on the task. As such, rewards are task specific, need prior specification, and might be sparse, i.e. limited to the agent `did not fail' \cite{sutton18}. At each time step, \emph{t}, the agent receives a state vector, $s_t$, describing what the agent can `see' of the task. The agent suggests an action, $a_t$, and returns this to the task environment. The agent then receives a reward, $r_{t+1}$, as the environment transitions to the next state, $s_{t+1}$. Gradient methods for RL might then perform a credit assignment step. The process then iterates until a terminal state or some episode limit is encountered. Only once the terminal state or episode limit is encountered will non-gradient methods, such as evolutionary computation (EC), define policy performance in terms of the reward accumulated over the entire episode.

EC methods such as CMA-ES and NES have recently been able to demonstrate that monolithic \emph{linear} policies can be discovered for real-valued control tasks that are frequently better than non-linear evolutionary methods, such as genetic programming (GP), see Table \ref{tab:related_performance} for some recent results.\footnote{Caveats might include that although the tasks have common definitions the version of the simulation environment and evaluation vary.} Conversely, although GP solutions might well be non-linear (which need not make them less interpretable), GP does not assume that all features in the state vector have to appear in a policy.\footnote{Both CMA-ES and NES assume that a monolithic linear model employs all features, thus solutions always consist of \# feature $\times$ \# action parameters.} This provides GP with the ability to associate certain features with different actions. In addition, GP may also provide the opportunity to discover how to decompose a task, such that different `modules' are applied repeatedly for the same context. Such decompositions might aid in the interpretability or ability to scale a policy to more difficult tasks. To date, MAPLE appears to be the only GP framework that uses modularity successfully on real-valued control tasks (Table \ref{tab:related_performance}.). 

\begin{table}[t]
\centering
\caption{Summary of comparator method results on benchmark RL tasks. Values for MAPLE, CGP and xNES+ had to be estimated visually from figures. CMA-ES, SIRL and ARS are the average test performance quoted from the original works. \dag indicates a result from ARS v1 versus v2. CMA-ES results are quoted as the best from 6 different versions \cite{wong27}}\label{tab:related_performance}
    {\scriptsize\begin{tabular}{lcccccc}
\toprule
Task & MAPLE \cite{vacher25} & CMA-ES \cite{wong27} & SIRL \cite{custode25} & xNES+ \cite{kimuraO24} & CGP \cite{nadizar27} & ARS \cite{wong27} \\
\midrule
\multicolumn{7}{l}{\emph{MuJoCo}} \\
Hopper & $3000-3400$ & $3021$ & $1026$ & $\approx3250$ & $\approx1150$ & $2292$\\
Walker2d & $2500-3000$ & $2575$ & $982$ & $\approx6200$ & $\approx1600$ & $2288$\\
HalfCheetah & $2500-3000$ & $4764$ & -- & $\approx6000$ & $\approx6200$ & $4297$\\
Swimmer & -- & $296$ & $223$ & $\approx350$ & $\approx280$ & $345$\dag \\
Reacher & -- & -- & $-8.71$ & $\approx-10$ & -- & --\\
Ant & $2000-3000$ & $2886$ & -- &  $\approx5500$ & $\approx1600$ & 2457\\
Humanoid & $4000$ & $808$ & -- & $\approx9500$ & -- & $1574$\\
\midrule
\multicolumn{6}{l}{\emph{Classic control}} \\
CartPole & -- & $500$ & -- & -- & $\approx500$ & $500$\\
MountainCar & -- & -- & -- & -- & $\approx-100$ & --\\
Acrobot & -- & $-72$ & -- & -- & $\approx-72$ & $-70$\\
\midrule
\multicolumn{6}{l}{\emph{Box2D}} \\
LunarLander & -- & $273$ & $272$ & -- & $\approx280$ & $264$\\
BipedalWalker & -- & $225$ & -- & -- & -- & $105$\\
\bottomrule
\end{tabular}}

\end{table}

In addition, computational cost or rather sample efficiency remains an issue of impact to any EC approach applied to RL tasks. Gradient methods (in RL tasks) have made extensive use of off-policy formulations in which each interaction with the task is recorded in a replay buffer, e.g. \cite{mnih15}. Thus, a training epoch implies that multiple updates are performed using previously visited states providing improved sample efficiency. However, as noted above, GP is evaluated episodically, thus additional limitations might well appear with agents repeating negative behaviours for an entire episode, leading to poor sample efficiency (e.g. see the commentary in \cite{nadizar27}). Recently, Custode and Iacca attempt to improve sample efficiency through a `social' model (SIRL in Table \ref{tab:related_performance}) in which all the individuals comprising the population experience the state under evaluation \cite{custode25}. A voting model identifies the single action deployed. Only during a latter phase of evolution are individuals independently evolved.

Offline reinforcement learning potentially maximizes sample efficiency by having an oracle (human or otherwise) perform the task and therefore construct a dataset (aka an idealized replay buffer). However, offline learning suffers from distributional drift that appears to preclude the naive application of symbolic regression methods \cite{macinnis27}. Earlier researchers showed that the combination of offline and online evolution may lead to more efficient solutions (e.g. \cite{dhebar24,medvet24}), but performance on real-valued action tasks in particular has been limited. Table \ref{tab:related_performance} provides a summary of current state-of-the-art in EC methods as applied to several popular discrete and real-valued action benchmarks.

Our approach will be to assume a Deep Reinforcement Learning (DRL) or a human oracle and then perform several phases of model building from the resulting offline dataset. The underlying motivation is to develop a decision list that defines how a set of linear actions are applied. Such actions represent \textbf{motion primitives} that the decision list chooses between. Thus, rather than assuming a single monolithic model we aim to provide a high-level description of the policy through the decision list. The decision list and initial motion primitives are identified offline given an original behaviour (demonstrations) defined by an oracle. The motion primitives are identified using a Expectation Maximization algorithm whereas GP constructs the decision list. Finally, the method of Augmented Random Search (ARS) \cite{mania18} is assumed to fine-tune any constants in the decision list and linear models. This last process is performed online, i.e. through interaction with the task. In summary, we assume
\begin{enumerate}
    \item \textbf{offline} evolution to identify the motion primitives (set of linear actions) from which a high level interpretable policy is defined in terms of a decision list;
    \item \textbf{online} fine tuning of the constants in the decision list and linear models using a simple stochastic optimizer.
\end{enumerate}

The resulting empirical study over 20 benchmark reinforcement learning tasks demonstrates that the proposed framework: 1) Typically provides state-of-the-art solutions (relative to recent comparator EC methods, Table \ref{tab:related_performance}). 2) Discovers a decision list policy that is interpretable for difficult RL tasks such as Humanoid. 3) Discovers policies that perform better than the original oracle in 11 of the 20 tasks.

\section{Proposed Framework}

\begin{figure}
    \centering
    \resizebox{\linewidth}{!}{
\definecolor{stagecolor}{RGB}{147,32,146}
\begin{tikzpicture}[node distance=7mm,
    every node/.append style={execute at begin node={\hyphenpenalty10000\relax}},
    stage/.style={draw, thick, rounded corners=2pt, draw=stagecolor, fill=stagecolor!12, text width=26mm, minimum height=21mm, align=center,
                  font=\small},
    data/.style={draw, thick, cylinder, shape border rotate=90, aspect=0.18, fill=white, text width=25mm,
                 minimum height=21mm, align=center, font=\small},
    env/.style={draw, thick, fill=white, text width=26mm, minimum height=9mm, align=center, font=\small},
    flow/.style={-{Latex[length=2.2mm]}, thick},
    phase/.style={draw, dashed, thick, black!55, rounded corners=4pt, inner sep=3mm},
    phaselabel/.style={font=\small\itshape, text=black!65, anchor=south west, inner sep=1pt}]
  \node[data] (data) {Oracle dataset\\(demonstrations)};
  \node[stage, right=of data] (motions) {Motion discovery\\(expectation-\\maximization)};
  \node[stage, text width=32mm, right=of motions] (evolve) {Decision list evolution\\(imitation learning)};
  \node[stage, text width=36mm, right=10mm of evolve] (tune) {Joint tuning of\\thresholds, expressions\\and linear models\\(ARS)};
  \node[env, text width=36mm, below=8mm of tune] (task) {Task environment};

  \draw[flow] (data) -- (motions);
  \draw[flow] (motions) -- (evolve);
  \draw[flow] (evolve) -- (tune);
  \draw[flow] ([xshift=-5mm]tune.south) -- node[left, font=\footnotesize] {actions} ([xshift=-5mm]task.north);
  \draw[flow] ([xshift=5mm]task.north) -- node[right, font=\footnotesize] {rewards} ([xshift=5mm]tune.south);

  \node[phase, fit=(data) (motions) (evolve)] (offline) {};
  \node[phase, fit=(tune) (task) (tune.north |- data.north)] (online) {};
  \node[phaselabel] at (offline.north west) {Offline};
  \node[phaselabel] at (online.north west) {Online};
\end{tikzpicture}
    }
    \caption{Overall framework for online/offline evolution of a decision list with (linear) motion primitives. Only the ARS tuning step is an online process}
    \label{fig:framework}
\end{figure}
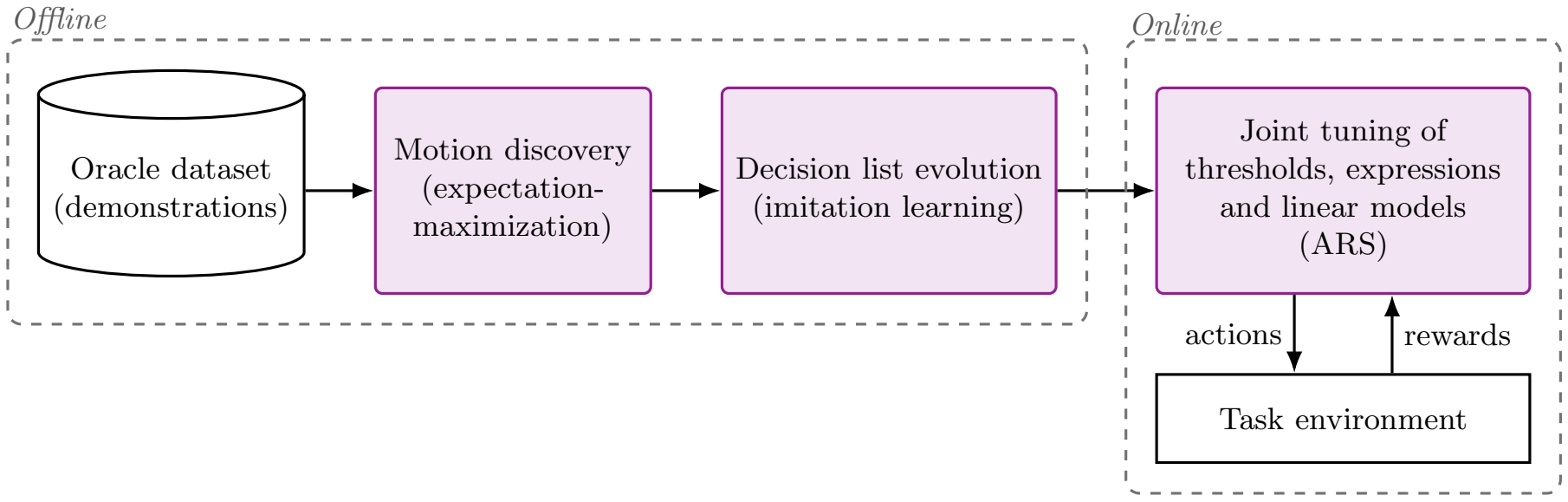

Our proposed framework for online/offline evolution of a decision list with motion primitives is summarized in Figure \ref{fig:framework} and detailed in the following subsections.

\subsection{Finding Motion Primitives for Continuous Control}

For continuous-action RL tasks, we adopt an offline RL starting point (e.g. \cite{minari,emmons22,prudencio24}), which is to say, we assume that an oracle is available that already demonstrates satisfactory performance on the target task.\footnote{For example, a Deep RL method or human expert.} For the discrete-action RL tasks, this element can be skipped. The oracle represents the source of state-action-reward tuples, $\langle s_t, a_t, r_{t+1} \rangle$, from which our goal is to discover a set of motion primitives (linear models) that describe the variety of behaviours that make up an oracle's actions. Earlier research has noted that attempting to directly fit a symbolic regression model to such data typically fails \cite{macinnis27}. Conversely, fitting the same data using a multi-layer perceptron (of lower complexity than the original oracle) often results in a successful policy.

With the oracle demonstrations in place, we apply an Expectation-Maximization (EM) algorithm to discretize the actions and introduce an optional dynamic programming step in said approach. This optional step penalizes switching the motion primitives between time steps. Our hypothesis is that the motion primitive chosen should remain in use for a consecutive sequence of time steps after. We later benchmark both cases independently.

\begin{figure}[htbp]
    \centering
    \resizebox{\linewidth}{!}{%
\definecolor{switchcolor}{RGB}{255,0,0}
\begin{tikzpicture}[x=12mm, y=7.5mm,
    cell/.style={draw=black!35, minimum width=8mm, minimum height=6.5mm, inner sep=0pt, font=\small},
    chosen/.style={cell, draw=black, fill=black!12, thick},
    stay/.style={line width=1.2pt, black},
    switch/.style={line width=1.2pt, switchcolor},
    lbl/.style={font=\small}]
  \foreach \dx/\route/\title/\costline/\totalline in {
      0/{1,3,1,3,3,3}/{Lowest cost per step}/{cost $6$, $3$ switches}/{total $6 + 3\lambda = 12$},
      8.2/{1,1,1,3,3,3}/{Lowest cost with switching penalty $\lambda = 2$}/{cost $8$, $1$ switch}/{total $8 + \lambda = 10$}} {
    \node[lbl, font=\normalsize] at (\dx + 3.5, 0.15) {\title};
    \foreach \t/\ca/\cb/\cc in {1/1/2/3, 2/3/2/1, 3/1/2/4, 4/3/2/1, 5/4/2/1, 6/4/3/1} {
      \node[cell] at (\dx + \t, -1) {$\ca$};
      \node[cell] at (\dx + \t, -2) {$\cb$};
      \node[cell] at (\dx + \t, -3) {$\cc$};
      \node[lbl, text=black!60] at (\dx + \t, -3.8) {$\t$};
    }
    \foreach \m in {1, 2, 3} {\node[lbl, anchor=east] at (\dx + 0.6, -\m) {motion $\m$};}
    \node[lbl, text=black!60, anchor=east] at (\dx + 0.6, -3.8) {step};
    \foreach \m [count=\t] in \route {
      \pgfmathtruncatemacro{\c}{{{1,2,3},{3,2,1},{1,2,4},{3,2,1},{4,2,1},{4,3,1}}[\t-1][\m-1]}
      \node[chosen] at (\dx + \t, -\m) {$\c$};
    }
    \foreach \m [count=\t, remember=\m as \prev (initially 0)] in \route {
      \ifnum\t>1
        \pgfmathtruncatemacro{\s}{\t - 1}
        \ifnum\m=\prev
          \draw[stay] (\dx + \s + 0.34, -\m) -- (\dx + \t - 0.34, -\m);
        \else
          \draw[switch, -latex] (\dx + \s + 0.34, -\prev) -- (\dx + \t - 0.34, -\m);
        \fi
      \fi
    }
    \node[lbl] at (\dx + 3.5, -4.6) {\costline};
    \node[lbl] at (\dx + 3.5, -5.3) {\totalline};
  }
\end{tikzpicture}
    }
    \caption{An illustrated example of the optional dynamic-programming step when the switching penalty is used to reduce the amount of switching between motion primitives despite incurring slightly greater cost without the penalty.}
    \label{fig:switching_penalty}
\end{figure}
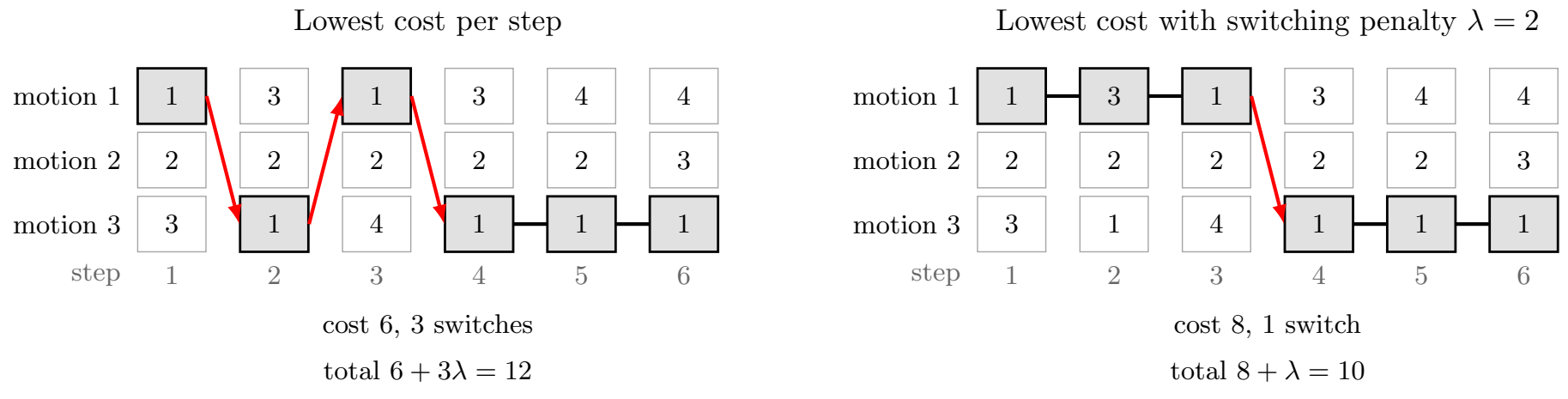

The EM algorithm is detailed as follows. A fitting phase and a re-labelling phase run iteratively either until the motion primitives converge or for a maximum number of iterations. To generate the coarse \textbf{initial set} of labels, the K-means clustering algorithm is applied to the actions; giving in turn, K clusters of continuous actions. With the coarse labels in place, the EM algorithm is seeded and can begin the fitting phase. In the \textbf{fitting phase}, ridge regression fits a linear model per cluster which minimizes the error on the clustered continuous actions. If fewer than the minimum amount of data points are assigned to a linear model, the linear model is instead fit to a random sample of data points.\footnote{Parameters for this are described further in Table \ref{tab:discretization_parameters} in Supplementary Material.} In the \textbf{re-labelling phase}, we present two choices: \begin{enumerate}
    \item re-label the actions such that they belong to the linear model with the least cost.
    \item introduce a switching penalty and assume a dynamic-programming approach, Figure \ref{fig:switching_penalty}.
\end{enumerate} A cost-table is computed prior to re-labelling each linear model. Assuming no switching penalty is applied, each cell's value ($C_{ij}$) is the squared Euclidean distance between the best linear model $i$ at that timestep $j$ with the oracle's action in the dataset. When a switching penalty is applied, the cost table's cell $C_{ij}$ is the smaller combined cost of \begin{enumerate}
    \item the current cost of the path without switching linear models plus the cost at this timestep.
    \item the minimum cost of all paths preceding this timestep plus a switching penalty $\lambda$.
\end{enumerate} Each cell keeps track of the previous path chosen through a pointer. When the final timestep is computed, the pointers are followed backwards from the lowest cost cell to determine the labels at each timestep.

The EM algorithm as described above is repeated from all values from 1 to $K$, and the value of K which reduces $X\%$\footnote{90\% is used in this work and is detailed in Table \ref{tab:discretization_parameters} in the supplementary material.} of the error otherwise covered by all $K$ motion primitives is chosen. This reduced K-value is a diminishing returns trade-off between error and the number of motion primitives used.

\subsection{Evolving a Decision List Policy by Imitation Learning}

Our policy representation is based on decision lists. At the highest level a decision list is a set of {\ttfamily if-elif-else} rules. Each rule makes a comparison between a variable number of predicates {\ttfamily pred\#}. The predicates themselves deploy comparisons on {\ttfamily expressions}. Finally, if a rule is true then the linear model or {\ttfamily motion} is deployed. Figure \ref{fig:policy-summary} illustrates some of the resulting policies.

A canonical GP algorithm (Algorithm \ref{alg:gp}) with parameters (Table \ref {tab:gp_config}, supplementary material) is used under an imitation learning objective. Thus, fitness of an individual under a discrete task is defined as the accuracy of the predicted action matching the oracle's action. Under continuous-action tasks is defined such that a value of 0 suggests that the oracle and the decision list incur the same cost and a value below 0 indicates that the decision list incurred more cost. Additionally, a small penalty is applied per rule and per predicate to encourage parsimony when the incurred cost is otherwise the same.

The definition of fitness for continuous and discrete actions follows:

 \begin{equation*}\label{eq:fitness}
  F =
  \begin{cases}
    \dfrac{1}{|\mathcal{B}|} \displaystyle\sum_{t \in \mathcal{B}}
      \bigl( C_t(\text{oracle}_t) - C_t(\text{individual}_t) \bigr) - \lambda \bigl( R(\pi) + P(\pi) \bigr)
      & \text{continuous,} \\[3ex]
    \dfrac{1}{|\mathcal{B}|} \displaystyle\sum_{t \in \mathcal{B}}
      \mathbb{I}\bigl[ \text{individual}_t = \text{oracle}_t \bigr] - \lambda \bigl( R(\pi) + P(\pi) \bigr)
      & \text{discrete,}
  \end{cases}
\end{equation*}
\noindent where $C_t$ denotes the cost of choosing the action compared to the original action in the dataset, $\text{oracle}_t$ and $\text{individual}_t$ represent the oracle's and the individual's choice at time step $t$, $R(\pi)$ the number of rules, $P(\pi)$ the number of predicates, and $\lambda$ the complexity penalty.
  
When selecting the next population, mutation or crossover is performed against the parent population by changing the rules or predicates randomly or swapping the rules and predicates randomly between two parents. A more complete description of variation is given in Table \ref{tab:variation} in the supplementary material.

\begin{algorithm}[t]
\caption{Genetic Programming Algorithm}\label{alg:gp}
\label{alg:gp}
\begin{algorithmic}[1]
\Require Population size $P$, generations $G$, elite count $N$
\State $\mathcal{P} \gets \operatorname{Random Population}()$
\For{$g = 1, \ldots, G$}
    \State $\mathbf{f} \gets \operatorname{Evaluate}(\mathcal{P})$
    \State $\mathcal{P}' \gets \operatorname{Elites}(\mathcal{P}, \mathbf{f}, N)$
    \While{$|\mathcal{P}'| < P$}
        \If{$\operatorname{Crossover}$}
            \State $p_1 \gets \operatorname{TournamentSelect}(\mathcal{P}, \mathbf{f})$
            \State $p_2 \gets \operatorname{TournamentSelect}(\mathcal{P}, \mathbf{f})$
            \State $o \gets \operatorname{Crossover}(p_1, p_2)$
        \Else
            \State $p \gets \operatorname{TournamentSelect}(\mathcal{P}, \mathbf{f})$
            \State $o \gets \operatorname{Mutate}(p)$
        \EndIf
        \State $\mathcal{P}' \gets \mathcal{P}' \cup \{o\}$
    \EndWhile
    \State $\mathcal{P} \gets \mathcal{P}'$
\EndFor
\end{algorithmic}
\end{algorithm}

\subsection{Pruning and Final Re-fitting of the Policy}

Prior to tuning online with ARS, unused rules and predicates are pruned from the champion policy. A data subset is sampled from the oracle dataset to verify that the pruned decision lists matches the original policy. Candidates for pruning the policy are generated by removing a rule or removing a predicate from the decision list. When all candidates are made, the first candidate that can remove a rule or predicate without changing the behaviour is selected. That pruned policy becomes the new policy and the process is repeated until no further edits can be made without introducing disagreement. Finally, the complete oracle dataset is then routed to the decision list for a final fitting of the linear models based on the data that the linear model or motion primitives actually receive after routing.

\subsection{Online Augmented Random Search Tuning}

With the champion policy in place, online policy optimization is performed. The thresholds across all predicates, the random constants in all expressions and the weights and biases in the linear models are flattened into a vector and jointly tuned by ARS \cite{mania18}. The parameters used for ARS are detailed in Table \ref{tab:ars} in supplementary material.

\section{Experimental Setup}

Table \ref{tab:oracle_datasets} summarizes the properties of the 20 tasks employed for this benchmarking study and the nature of the oracle assumed to provide the offline RL dataset. For the most part we focus on tasks with real-valued actions and dense rewards. The 8 tasks from the MuJoCo suite have seen widespread use by comparator methods (Table \ref{tab:related_performance}). The principle caveat is that the simulator has been updated to correct for errors; our evaluation reports on v5 for each task\footnote{The Ant-v5 task has been updated to include more state features than previously given in Ant-v4, however, we make use of only the features in the Ant-v4 state vector to match the oracle dataset.}. The Bipedal Walker (real-valued action) task is also benchmarked but uses the box2d physics engine. In addition, 4 discrete action tasks are included to verify the generality of the approach (Acrobot, Cart Pole, Mountain Car, Lunar Lander). Two additional task domains are then considered, where these describe robotics tasks in higher dimensions: Franka Kitchen \cite{gupta19,fu20} and Adroit Hand \cite{rajeswaran18,fu20}.

\begin{table}
\centering
\caption{Summary of oracle datasets. Oracle denotes the source of the offline dataset. Rows is the cardinality for the offline dataset. $|$State$|$ and $|$Action$|$ are the dimension of the state (input) and action (output) respectively}
\label{tab:oracle_datasets}
    {\scriptsize\begin{tabular}{lcccclcccc}
\toprule
Task & Oracle & Rows & $|$State$|$ & $|$Action$|$ & Task & Oracle & Rows & $|$State$|$ & $|$Action$|$\\
\midrule
\multicolumn{5}{l}{\emph{MuJoCo}} & \multicolumn{5}{l}{\emph{Box2D}} \\
Hopper & TD3 & $199{,}535$ & $11$ & $3$ & LunarLander$^{*}$ & PPO & $39{,}756$ & $8$ & $4$\\
Walker2d & TD3 & $200{,}000$ & $17$ & $6$ & BipedalWalker & TD3 & $199{,}976$ & $24$ & $4$\\
\cmidrule{6-10}
HalfCheetah & TD3 & $200{,}000$ & $17$ & $6$ & \multicolumn{5}{l}{\emph{Adroit}}\\
Swimmer & CMA-ES & $200{,}000$ & $8$ & $2$ & Door & DAPG & $199{,}852$ & $39$ & $28$\\
Reacher & TD3 & $50{,}000$ & $10$ & $2$ & Hammer & DAPG & $199{,}867$ & $46$ & $26$\\
Pusher & SAC & $100{,}000$ & $23$ & $7$ & Pen & DAPG & $199{,}954$ & $45$ & $24$\\
Ant & SAC & $30{,}000$ & $27$ & $8$ & Relocate & DAPG & $199{,}817$ & $39$ & $30$\\
Humanoid & TQC & $30{,}000$ & $376$ & $17$\\
\midrule
\multicolumn{5}{l}{\emph{Classic control}}  & \multicolumn{5}{l}{\emph{Franka Kitchen}}\\
CartPole & PPO & $199{,}572$ & $6$ & $2$ & Kitchen & Human & $199{,}865$ & $59$ & $9$\\
MountainCar & CMA-ES & $199{,}970$ & $2$ & $3$ & Kitchen-mixed & Human & $199{,}865$ & $59$ & $9$\\
Acrobot & PPO & $70{,}682$ & $6$ & $3$ & Top burner & Human & $199{,}865$ & $59$ & $9$\\
\bottomrule
\multicolumn{3}{l}{\rlap{\scriptsize $^{*}$Balanced: resampled to equal counts of its four actions.}} \\
\end{tabular}}
\end{table}

The Franka Kitchen benchmark is based on tasks that involve manipulating common household items in a kitchen environment using a 9 degrees of freedom (Franka) robot arm. The tasks are described in terms of a 59 dimensional state space and 9 real valued actions under sparse rewards (zero unless the task is completed). Franka Kitchen in the following experiments has an additional reward-shaping term applied such that a progress bonus gives a fractional reward based on partial progress to the next goal. The reward shaping is not applied at evaluation time. The Adroit Hand environment is based on a `shadow dexterous hand' that supports up to 30 degrees of freedom and rewards are again sparse. Four tasks are described: Door (hand has to open a door latch), Hammer (hand has to hammer a nail into a board), Pen (manipulate a pen into a goal position/ rotation), and Relocate (hand picks up a ball and positions it at a target location. Both the Franka and Adroit environments assume human demonstrations to define the oracle behaviours, thus more limited volume of data than under tasks with a DRL oracle.

For each task, an oracle has to be identified. Rather than assume that a single oracle will always be appropriate, we assume a cross section of different candidate oracles, Table \ref{tab:oracle_complexities}. Thus, we assume CMA-ES for Swimmer and Mountain Car where these tasks might trip up DRL agents. TD3, SAC and PPO represent widely applied DRL approaches for tasks with real-valued actions. TQC \cite{DBLP:journals/corr/abs-2005-04269} represents a more recent DRL approach that scales better on the high dimensional Humanoid task. Both Kitchen and Adroit assume human oracles to some degree.

\begin{table}[htbp]
    \centering
        \caption{Complexity of the oracle policies}
    \label{tab:oracle_complexities}
{\scriptsize\begin{tabular}{llr}
\toprule
Oracle & Tasks & Parameters \\
\midrule
TD3 \cite{fujimoto18} & Hopper, Walker2d, & $\approx 125$k--$132$k \\
    & HalfCheetah, Reacher, & \\
    & BipedalWalker & \\
SAC \cite{haarnoja18} & Pusher, Ant & $\approx 74$k--$75$k \\
TQC \cite{DBLP:journals/corr/abs-2005-04269} & Humanoid & $\approx 160$k \\
PPO \cite{schulman17} & CartPole, LunarLander, & $\approx 4.6$k--$5.0$k \\
    & Acrobot & \\
CMA-ES \cite{wong27} & Swimmer, MountainCar & $16$--$30$ \\
DAPG and humans$^{\dagger}$ \cite{fu20}& Door, Hammer, & $\approx 3.3$k--$8.7$k \\
    & Pen, Relocate & \\
Humans \cite{fu20} & Kitchen, Kitchen-mixed, & --- \\
    & Top burner & \\
\bottomrule
\multicolumn{3}{l}{\scriptsize $^{\dagger}$The data also holds human demonstrations, which no parameter count describes.} \\
\end{tabular}}

\end{table}

\section{Results}

Performance is expressed post-training as the cumulative task reward over the 10 champions from each training run. Each post-training evaluation is performed over 100 unseen episodes, Table \ref{tab:our_results} quotes the original average performance of the oracle, the average performance of the policy following completion of all \textit{offline} training, and average performance of the policy following all steps of training (Offline + ARS).

The role of the optional dynamic programming step is ablated in Table \ref{tab:results-ablation}, in Figures \ref{fig:training-ablation}, \ref{fig:complexity-ablation} in the supplementary material. The performance of the final policies are largely the same regardless of whether the switching penalty is applied, however, there exists a trade-off between sample-efficiency and complexity. When a switching penalty is applied, the final policies have less complexity, but when a switching penalty is not applied, the sample efficiency is improved on some tasks (e.g. Hopper and HalfCheetah) but slower on others (e.g. Reacher).

It is clear that under the Classic control and Adroit benchmarks, offline training alone is sufficient to provide performance that approaches that of the Oracle. However, the Offline policies never approach that of the Oracle under the MuJoCo and Franka Kitchen domains. \emph{Online} optimization of the policies identified offline using ARS results in solutions that exceed or equal the performance of the Oracle for 11 of the 20 tasks and marginally less for another 4 (Hopper, Reacher, Pusher, Kitchen).

\begin{table}[t]
    \centering
        \caption{Performance of proposed framework. `Oracle' is the performance of the DL used to provide the data for offline training. Online and Online+ARS reflect the proposed framework without/with ARS optimization}
    \label{tab:our_results}
    {\scriptsize\begin{tabular}{lrrr}
\toprule
Task & Oracle & Offline & Offline + ARS \\
\midrule
\multicolumn{4}{l}{\emph{MuJoCo}} \\
Hopper & $3{,}588$ & $606.6 \pm 435.6$ & $3{,}559 \pm 169$ \\
Walker2d & $4{,}642$ & $439.7 \pm 152.2$ & $5{,}134 \pm 99$ \\
HalfCheetah & $6{,}536$ & $-100.9 \pm 236.0$ & $7{,}729 \pm 929$ \\
Swimmer & $354.6$ & $280.3 \pm 33.3$ & $362.1 \pm 0.4$ \\
Reacher & $-4.96$ & $-5.51 \pm 0.16$ & $-5.10 \pm 0.30$ \\
Pusher & $-31.2$ & $-33.1 \pm 0.2$ & $-32.8 \pm 0.2$ \\
Ant & $6{,}719$ & $49.4 \pm 435.9$ & $5{,}227 \pm 251$ \\
Humanoid & $10{,}510$ & $365.2 \pm 99.0$ & $9{,}181 \pm 626$ \\
\midrule
\multicolumn{4}{l}{\emph{Classic control}} \\
CartPole & $500.0$ & $500.0 \pm 0.0$ & $500.0 \pm 0.0$ \\
MountainCar & $-98.4$ & $-100.0 \pm 2.5$ & $-98.7 \pm 0.2$ \\
Acrobot & $-69.0$ & $-88.3 \pm 5.3$ & $-81.3 \pm 5.8$ \\
\midrule
\multicolumn{4}{l}{\emph{Box2D}} \\
LunarLander & $259.5$ & $164.9 \pm 45.5$ & $234.8 \pm 21.5$ \\
BipedalWalker & $316.8$ & $-1.29 \pm 60.40$ & $341.0 \pm 3.0$ \\
\midrule
\multicolumn{4}{l}{\emph{Adroit}} \\
Door & $3{,}041$ & $2{,}992 \pm 7$ & $3{,}321 \pm 32$ \\
Hammer & $16{,}359$ & $16{,}145 \pm 17$ & $18{,}574 \pm 85$ \\
Pen & $3{,}371$ & $2{,}455 \pm 220$ & $4{,}482 \pm 138$ \\
Relocate & $4{,}413$ & $3{,}225 \pm 389$ & $4{,}952 \pm 68$ \\
\midrule
\multicolumn{4}{l}{\emph{Franka Kitchen}} \\
Kitchen & $3.73$ & $0.07 \pm 0.15$ & $3.56 \pm 0.45$ \\
Kitchen-mixed & $3.48$ & $0.02 \pm 0.05$ & $3.87 \pm 0.15$ \\
Top burner & $0.97$ & $0.00 \pm 0.00$ & $0.98 \pm 0.02$ \\
\bottomrule
\end{tabular}}

\end{table}

\begin{figure}[htbp]
    \centering
    \resizebox{\linewidth}{!}{%
        \input{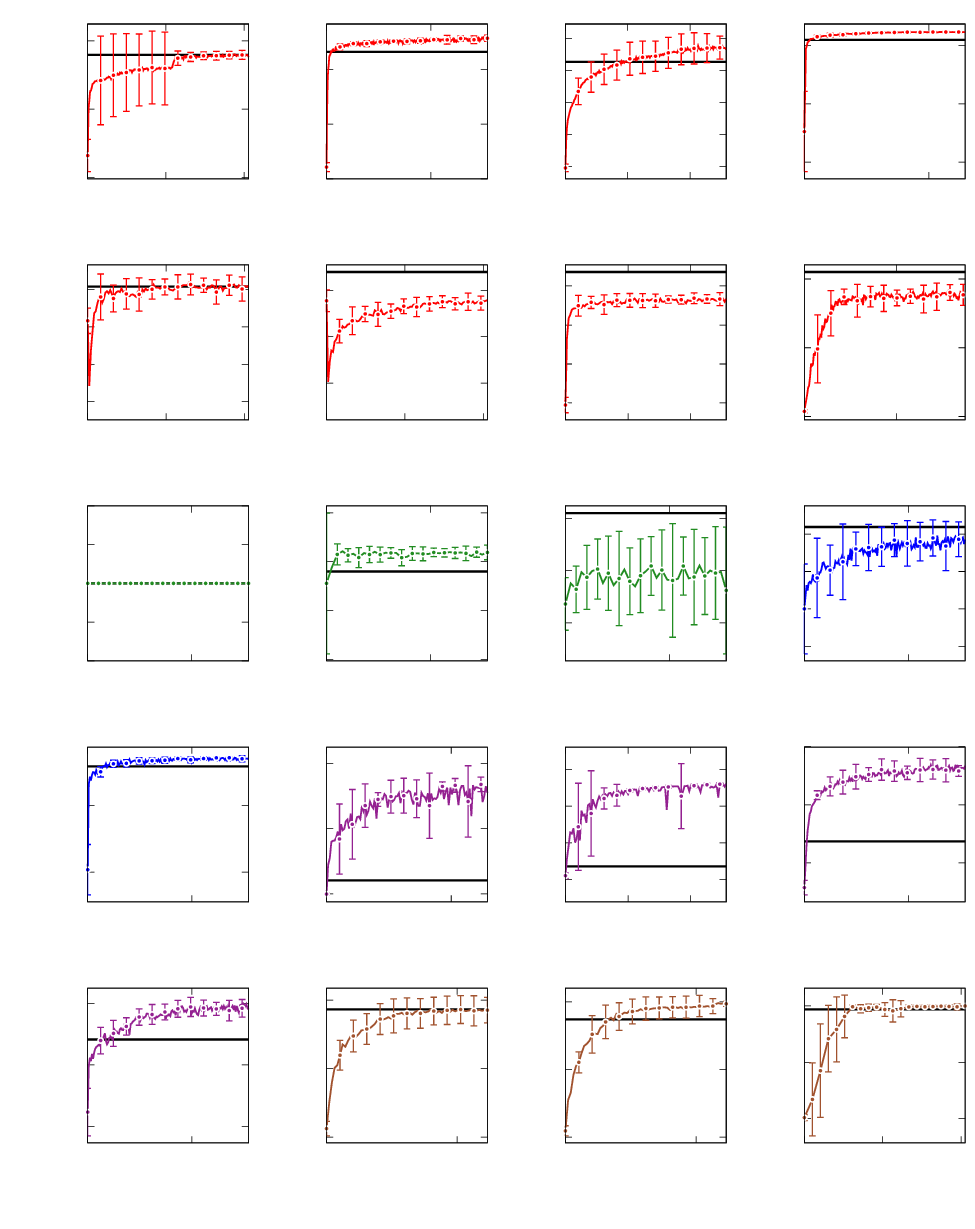}
    }
    \caption{Test performance as a function of training episodes.}
    \label{fig:training}
\end{figure}

Figure \ref{fig:training} illustrates how the ARS online training improves the performance of the respective decision list policies. Acrobot and Lunar Lander demonstrate lower rates of improvement, most likely due to the stochastic nature of the task; so they might benefit from more evaluations per fitness assessment. That said, Hopper is also particularly noisy, but goes through a distinct phase change after which the policy is both stable and very similar to that of the oracle. Ant and Humanoid did not reach the performance of the oracle, but exceed (or comparable) to all the comparator evolutionary algorithms, Table \ref{tab:related_performance}.

\begin{figure}[htbp]
    \centering
    \resizebox{\linewidth}{!}{%
        \input{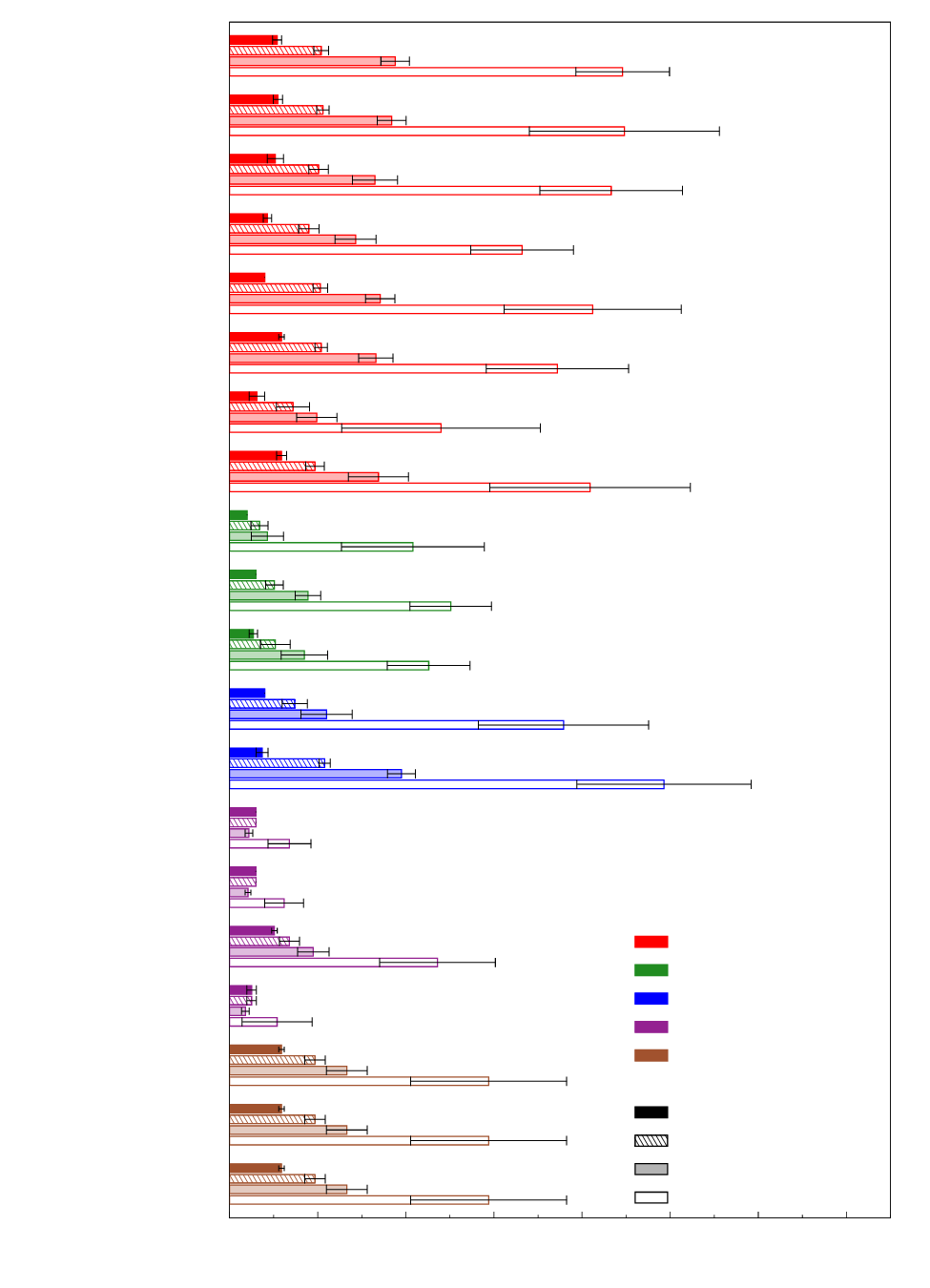}
    }
    \caption{Counts of primitive present in each policy as averaged over 10 trials}
    \label{fig:complexity}
\end{figure}

\lstdefinelanguage{policy}{sensitive=true, alsoletter={-:}, alsodigit={.},
  morekeywords=[1]{:accelerate-left,:accelerate-right,:coast,:do-nothing,:fire-left-engine,:fire-main-engine,
    :fire-right-engine,:motion-1,:motion-2,:motion-3,:motion-4,:motion-5,:motion-6,:negative-torque,:no-torque,
    :positive-torque,:push-left,:push-right},
  morekeywords=[2]{define,defpolicy,if,elif,else},
  morekeywords=[3]{acrobot,adroit-door,adroit-hammer,adroit-pen,adroit-relocate,ant,bipedal-walker,cartpole,
    half-cheetah,hopper,humanoid,kitchen,kitchen-mixed,kitchen-top-burner,lunar-lander,mountain-car,pusher,reacher,
    swimmer,walker2d}}
\lstdefinestyle{decisionlist}{language=policy, basicstyle=\ttfamily\scriptsize, columns=fullflexible,
  keepspaces=true, breaklines=true, breakatwhitespace=true, postbreak=\mbox{\hspace{1em}},
  showstringspaces=false, frame=lines, framesep=3pt, xleftmargin=0pt, aboveskip=2pt, belowskip=2pt,
  keywordstyle=[1]\color[RGB]{72,61,139}, keywordstyle=[2]\color[RGB]{147,32,146},
  keywordstyle=[3]\color[RGB]{0,0,255}}
\lstdefinestyle{decisionlistsummary}{style=decisionlist, basicstyle=\ttfamily\scriptsize,
  aboveskip=24pt, belowskip=4pt}
\begin{figure}[p]
\centering
\begin{minipage}[t]{\linewidth}
\begin{lstlisting}[style=decisionlistsummary]
(define
  (pred1 (> (+ hammer-z (* 3.0166 nail-head-x)) 0.2518))
  (pred2 (< (sqrt (abs (* (* ffj0 mfj2) lfj0))) 0.6493)))

(defpolicy adroit-hammer
  (if pred1 :motion-2)
  (elif pred2 :motion-1)
  (else :motion-3))
\end{lstlisting}
{\footnotesize The policy for Adroit Hammer with a held-out mean across 100 episodes of $18{,}701$.\par}
\end{minipage}
\hfill
\begin{minipage}[t]{\linewidth}
\begin{lstlisting}[style=decisionlistsummary]
(define
  (pred1 (< (+ lfj4 (* 0.1620 pen-wx)) -0.4122))
  (pred2 (> (+ thj4 (+ mfj0 pen-dir-y)) 1.4221))
  (pred3 (< pen-minus-target-x 0.0317))
  (pred4 (< pen-dir-z 0.0087))
  (pred5 (< lfj0 0.9034))
  (pred6 (< (+ wrj0 (* 0.7290 ffj1)) -0.4420))
  (pred7 (< ffj1 0.9518)))

(defpolicy adroit-pen
  (if pred1 :motion-5)
  (elif (and pred2 pred3) :motion-2)
  (elif (and pred4 pred5) :motion-3)
  (elif pred6 :motion-5)
  (elif pred7 :motion-1)
  (else :motion-4))
\end{lstlisting}
{\footnotesize The policy for Adroit Pen with a held-out mean across 100 episodes of $4{,}672$.\par}
\end{minipage}

\medskip
\begin{minipage}[t]{\linewidth}
\begin{lstlisting}[style=decisionlistsummary]
(define
  (pred1 (> right-knee-angle -0.0897))
  (pred2 (> (norm torso-qx torso-qy torso-qz) 0.5622))
  (pred3 (> right-knee-angle 0.0135))
  (pred4 (> (+ (- left-elbow-angle (* 1.1655 left-shin-inertia-2)) (* 1.2888 right-thigh-inertia-8)) -0.4922))
  (pred5 (between (angle right-foot-inertia-8 left-foot-contact-force-0) -150.5126 -53.5966))
  (pred6 (< (- (- right-knee-angle (* 0.1132 (* right-shin-inertia-3 torso-height))) (* 0.0112 right-elbow-angle)) -1.0491))
  (pred7 (> right-thigh-comvel-4 0.2009))
  (pred8 (between (angle (* lwaist-inertia-8 left-hip-z-angle) (abs torso-inertia-2)) -153.7896 20.7418))
  (pred9 (> (- right-thigh-inertia-1 (* 6.9913 left-thigh-inertia-3)) 0.4672))
  (pred10 (< (- (- right-knee-angle (* 0.1144 pelvis-comvel-2)) (* 0.0107 left-knee-velocity)) -1.0005))
  (pred11 (< (- (* 0.0100 right-thigh-comvel-4) (p-div left-foot-contact-force-3 right-lower-arm-inertia-8)) -10.3673))
  (pred12 (> (+ (* right-hip-y-velocity -0.6912) right-lower-arm-inertia-2) 2.3075))
  (pred13 (< left-knee-velocity 18.6165)))

(defpolicy humanoid
  (if (and pred1 pred2) :motion-3)
  (elif (and pred3 (not pred4)) :motion-3)
  (elif (and (not pred5) pred6 pred7) :motion-1)
  (elif pred8 :motion-2)
  (elif pred9 :motion-5)
  (elif (and pred10 pred11) :motion-1)
  (elif pred12 :motion-2)
  (elif pred13 :motion-4)
  (else :motion-2))
\end{lstlisting}
{\footnotesize The policy for Humanoid with a held-out mean across 100 episodes of $9{,}976$.\par}
\end{minipage}

\medskip
\caption{The champion policies for Adroit Hammer, Adroit Pen, and Humanoid.}
\label{fig:policy-summary}
\end{figure}

We assume 4 count based metrics to summarize the complexity of policies under the decision list representation. Figure \ref{fig:complexity} summarizes the complexity of the solutions over all 20 tasks from multiple perspectives using these metrics. It is apparent that the Adroit and Kitchen tasks had simpler solutions whereas tasks from Box2D and MuJoCo were more complex. Generally, there were more operators than predicates, more predicates than rules and more rules than motions. Policies for Adroit Hammer, Door and Relocate were particularly open to interpretation. Figure \ref{fig:policy-summary} details policies for 3 tasks (supplementary material lists champion polices for all tasks).

\section{Discussion}

Despite the complexity of the Deep RL architectures, it is apparent that the policies learned by the oracles are not inherently complex in themselves. Some of this can be attributed to the need for online RL to discover a value function to solve a task whereas offline RL does not. Likewise, online RL methods such as CMA-ES, xNES+ and others have shown that only a single monolithic linear model is necessary for high episodic returns on some of these tasks; whereas results using GP illustrate that equally simple non-linear models might exist.

Our work illustrates that a piecewise linear solution is sufficient to imitate policies discovered by (online) oracles whether human, DRL or CMA-ES; at least for the benchmarks considered. The distinction we make is that a piecewise linear solution (versus a single monolithic solution) provides a narrative for how a linear policy is deployed. Such narratives provide the basis for interpretable solutions when viewed by, say, a domain expert.

One important caveat is that although offline RL through imitation learning is sufficient to identify the decision list or piecewise linear decomposition, it is currently not sufficient to `solve' the task. Online RL is still necessary to optimize the constants given the relevant decision list, in this case using ARS. Whether CMA-ES, xNES or something else represent a more efficient mechanism for parameter tuning remains a topic for future research. 

Similarly, we assume that there exists an oracle dataset of demonstrations in order to find the linear models. On the one hand, this means that advances in DRL can be directly used to develop better piecewise linear policies. On the other hand, this might simulate further research into evolving the decision lists and linear models jointly through online RL.

Finally, it is well understood in offline reinforcement learning literature that imitation learning is subject to problems such as distribution shift. However, there is a reward signal available from the dataset that we are not currently making use of. Future work might use this to mitigate distribution shift or to support sub-trajectory stitching.

\section{Conclusion}
The concept of imitation learning is revisited through three phases of policy development: 1) Offline construction of action / motion primitives from an oracle's data; 2) Offline identification of decision lists using GP to organize the motion primitives, providing a mechanism for interpreting the policy; and 3) Online tuning of the motion primitives and decision list constants using ARS. Benchmarking over 20 reinforcement learning tasks with real-valued / discrete actions and dense / sparse rewards indicates that the resulting policies typically exceed that from previous evolutionary approaches. Moreover, in the case of 11 of the benchmarks, the decision list policy exceeded the performance of the original oracle. Moreover, on tasks such as Ant and Humanoid, although we did not reach the performance of the oracle, we did exceed/match the state-of-the-art evolutionary methods.

Unlike the original DRL oracle we provide interpretable policies in which a decision list specifies the conditions under which linear motion primitives are deployed. Going forward, we note that the 3 components of the framework answer specific model building questions, but each component might be optimized further by adopting different algorithms.

\begin{credits}
\subsubsection{\ackname} The authors gratefully acknowledge support from NSERC Discovery program RGPIN-2025-06332.

\subsubsection{\discintname}
The authors have no competing interests to declare that are
relevant to the content of this article.
\end{credits}

\bibliographystyle{splncs04}
\bibliography{EuroGP_refs}

\section{Supplementary Material}

\begin{table}
\centering
\caption{Parameters for the Expectation Maximization algorithm used for discretizing the continuous actions into linear model motion primitives.}
\label{tab:discretization_parameters}
\begin{tabular}{ll}
\toprule
Parameter & Value \\
\midrule
\multicolumn{2}{l}{\emph{Number of motions}} \\
$K$ tried & $1$ to $8$ \\
Share of error reduction for choosing $K$ & $0.9$ \\
\midrule
\multicolumn{2}{l}{\emph{$k$-means initialization}} \\
Distance & squared Euclidean \\
Iterations & at most $20$ \\
\midrule
\multicolumn{2}{l}{\emph{Expectation-maximization}} \\
Ridge penalty & $10^{-3}$ \\
Switching penalty & $3$ \\
Iterations & at most $15$ \\
Fewest rows to refit a motion & $\lfloor (d+1)/4 \rfloor + 20$ \\
Rows sampled below that & $\max(200, \lfloor N/4K \rfloor)$ \\
\bottomrule
\end{tabular}

\end{table}

\begin{table}
\centering

\caption{Parameters for the Genetic Programming algorithm used for evolving the decision-list policies offline.}\label{tab:gp_config}
\begin{tabular}{ll}
\toprule
Parameter & Value \\
\midrule
\multicolumn{2}{l}{\emph{Search}} \\
Population size & $1{,}000$ \\
Generations & $200$ \\
Selection & tournament, size $5$ \\
Elitism & best $2\%$ ($20$ individuals) kept unchanged \\
$P(\text{mutation} / \text{crossover})$ & $0.7$ / $0.3$ (each child one or the other) \\
\midrule
\multicolumn{2}{l}{\emph{Fitness}} \\
Batch size & $40{,}000$ (one fixed random sample per search) \\
Complexity penalty, continuous actions & $0.002$ per rule and per predicate \\
Complexity penalty, discrete actions & $0.001$ per rule and per predicate \\
\midrule
\multicolumn{2}{l}{\emph{Policies}} \\
Rules & at most $10$, plus the default rule \\
Predicates per rule & at most $3$ \\
Predicates & $<$ and $>$ (against a threshold), \texttt{between} \\
 & \texttt{(between $\theta$ $a$ $b$)}: $\theta$ on the arc from $a$ \\
 & counterclockwise to $b$, in degrees \\
Unary operators & $|x|$, $\sqrt{|x|}$, $x^2$, $-x$ \\
Binary operators & $+$, $-$, $\times$, protected $\div$, \texttt{angle} (atan2 in degrees) \\
Expression tree depth & at most $5$ ($3$ when created) \\
\bottomrule
\end{tabular}

\end{table}

\begin{table}[t]
\centering
\caption{Types of variation (mutation and crossover) applied to individuals and their respective probabilities.}
\label{tab:variation}
{\scriptsize{\renewcommand{\arraystretch}{1.3}\setlength{\tabcolsep}{8pt}%
\begin{tabular}{l|p{7cm}|c}
\hline
Variation type & Description & Prob. \\
\hline
\textbf{Mutation} & & \\
\quad Threshold & A random predicate in a random rule has its threshold moved. Either the expression in the threshold is evaluated on a sample from the dataset and the new threshold is now the result of the expression \textbf{(p=0.3)} or the threshold is perturbed with Gaussian noise (0.2 $\times$ std of the result of the expression) \textbf{(p=0.7)}. & $3/13$ \\
\quad Expression & If the expression is a number, the number is perturbed by multiplying by a random factor $e^{N(0, 0.3)}$. With $\textbf{p=0.1}$, the sign also flips. If the expression is an operator, with $\textbf{p=0.4}$ the operator is swapped with another with the same arity, otherwise with $\textbf{p=0.6}$, the operator is replaced with a new sub-expression. If the expression is an observation, it is replaced with another sub-expression.& $1/13$ \\
\quad Replace predicate & A random predicate is replaced with a new predicate. & $1/13$ \\
\quad Negate & A predicate is wrapped with NOT such that when the old predicate would evaluate T/NIL, it would now evaluate NIL/T. & $1/13$ \\
\quad Add predicate & A new predicate is added to a random rule, nothing happens if the rule already has the maximum number of predicates. & $1/13$ \\
\quad Delete predicate & A random predicate is removed from a random rule. If the rule has only one predicate, nothing happens. & $1/13$ \\
\quad Add rule & A random rule is inserted at a random position in the decision list. If the policy already has the maximum number of rules, nothing happens. & $1/13$ \\
\quad Delete rule & A rule is randomly removed from the decision list. If that rule is the only rule other than the default rule, nothing happens. & $1/13$ \\
\quad Move rule & A rule is randomly taken out and reinserted at a random position in the decision list. Nothing happens if there is only one rule.& $1/13$ \\
\quad Rule motion & A random rule has a new motion assigned to it.& $1/13$ \\
\quad Default motion & The default rule has a new motion assigned to it.& $1/13$ \\
\hline
\textbf{Crossover} & & \\
\quad Rule & Two rule lists are spliced at a random point between two parents.& $1/2$ \\
\quad Predicate & With \textbf{p=0.3}, a predicate is added from parent B to parent A, otherwise a predicate in parent A is replaced with a predicate from parent B. & $3/10$ \\
\quad Subtree & Replace the expression within a predicate in parent A with another expression from parent B.& $1/5$ \\
\hline
\end{tabular}}}

\end{table}

\begin{table}[t]
\centering
\caption{Parameters for the Augmented Random Search (ARS) tuning of thresholds, constants in expressions, and linear models.}
\label{tab:ars}
{\scriptsize\begin{tabular}{lrrrrrr}
\toprule
Task & Iterations & Directions & Top & Step & Noise & Episodes \\
\midrule
\multicolumn{7}{l}{\emph{MuJoCo}} \\
Hopper & $2{,}000$ & $64$ & $16$ & $0.005$ & $0.0075$ & $4$ \\
Walker2d & $3{,}000$ & $64$ & $16$ & $0.01$ & $0.015$ & $4$ \\
HalfCheetah & $2{,}000$ & $64$ & $16$ & $0.02$ & $0.03$ & $2$ \\
Swimmer & $1{,}000$ & $64$ & $16$ & $0.02$ & $0.03$ & $2$ \\
Reacher & $1{,}000$ & $64$ & $16$ & $0.02$ & $0.03$ & $8$ \\
Pusher & $1{,}000$ & $64$ & $16$ & $0.02$ & $0.03$ & $8$ \\
Ant & $2{,}000$ & $64$ & $16$ & $0.01$ & $0.015$ & $4$ \\
Humanoid & $2{,}000$ & $64$ & $16$ & $0.002$ & $0.003$ & $4$ \\
\midrule
\multicolumn{7}{l}{\emph{Classic control}} \\
CartPole & $300$ & $32$ & $8$ & $0.02$ & $0.03$ & $4$ \\
MountainCar & $300$ & $32$ & $8$ & $0.02$ & $0.03$ & $4$ \\
Acrobot & $300$ & $32$ & $8$ & $0.02$ & $0.03$ & $4$ \\
\midrule
\multicolumn{7}{l}{\emph{Box2D}} \\
LunarLander & $1{,}500$ & $64$ & $16$ & $0.02$ & $0.03$ & $4$ \\
BipedalWalker & $1{,}500$ & $64$ & $16$ & $0.01$ & $0.015$ & $4$ \\
\midrule
\multicolumn{7}{l}{\emph{Adroit}} \\
Door & $1{,}000$ & $32$ & $8$ & $0.002$ & $0.004$ & $4$ \\
Hammer & $1{,}000$ & $64$ & $16$ & $0.002$ & $0.004$ & $4$ \\
Pen & $1{,}500$ & $32$ & $8$ & $0.002$ & $0.004$ & $8$ \\
Relocate & $1{,}500$ & $64$ & $16$ & $0.002$ & $0.004$ & $4$ \\
\midrule
\multicolumn{7}{l}{\emph{Franka Kitchen}} \\
Kitchen & $600$ & $64$ & $16$ & $0.006$ & $0.01$ & $8$ \\
Kitchen-mixed & $600$ & $64$ & $16$ & $0.006$ & $0.01$ & $8$ \\
Top burner & $600$ & $64$ & $16$ & $0.006$ & $0.01$ & $8$ \\
\bottomrule
\end{tabular}}

\end{table}

\begin{table}
\centering
\caption{Results after ablation, *indicates trials without a switching penalty while finding motion primitives}
\label{tab:results-ablation}
\begin{tabular}{lcccc}
\toprule
Task & \multicolumn{2}{l}{No switching penalty$^*$} & \multicolumn{2}{l}{With switching penalty} \\
 & Offline & Offline + ARS & Offline & Offline + ARS \\
\midrule
\multicolumn{5}{l}{\emph{MuJoCo}} \\
Hopper & $503.8 \pm 386.0$ & $3{,}600 \pm 68$ & $606.6 \pm 435.6$ & $3{,}559 \pm 169$ \\
Walker2d & $467.7 \pm 383.9$ & $5{,}096 \pm 118$ & $439.7 \pm 152.2$ & $5{,}134 \pm 99$ \\
HalfCheetah & $-190.1 \pm 285.0$ & $8{,}141 \pm 955$ & $-100.9 \pm 236.0$ & $7{,}729 \pm 929$ \\
Swimmer & $194.3 \pm 97.5$ & $362.2 \pm 0.9$ & $280.3 \pm 33.3$ & $362.1 \pm 0.4$ \\
Reacher & $-5.49 \pm 0.08$ & $-5.15 \pm 0.37$ & $-5.51 \pm 0.16$ & $-5.10 \pm 0.30$ \\
Pusher & $-33.2 \pm 0.6$ & $-32.9 \pm 0.3$ & $-33.1 \pm 0.2$ & $-32.8 \pm 0.2$ \\
Ant & $80.9 \pm 386.3$ & $5{,}305 \pm 274$ & $49.4 \pm 435.9$ & $5{,}227 \pm 251$ \\
Humanoid & $402.8 \pm 185.5$ & $9{,}428 \pm 315$ & $365.2 \pm 99.0$ & $9{,}181 \pm 626$ \\
\midrule
\multicolumn{5}{l}{\emph{Box2D}} \\
BipedalWalker & $13.5 \pm 63.7$ & $338.2 \pm 2.3$ & $-1.29 \pm 60.40$ & $341.0 \pm 3.0$ \\
\midrule
\multicolumn{5}{l}{\emph{Adroit}} \\
Door & $2{,}960 \pm 36$ & $3{,}327 \pm 34$ & $2{,}992 \pm 7$ & $3{,}321 \pm 32$ \\
Hammer & $16{,}087 \pm 165$ & $18{,}450 \pm 154$ & $16{,}145 \pm 17$ & $18{,}574 \pm 85$ \\
Pen & $2{,}226 \pm 116$ & $4{,}360 \pm 137$ & $2{,}455 \pm 220$ & $4{,}482 \pm 138$ \\
Relocate & $3{,}967 \pm 170$ & $4{,}903 \pm 130$ & $3{,}225 \pm 389$ & $4{,}952 \pm 68$ \\
\midrule
\multicolumn{5}{l}{\emph{Franka Kitchen}} \\
Kitchen & $0.01 \pm 0.02$ & $3.58 \pm 0.44$ & $0.07 \pm 0.15$ & $3.56 \pm 0.45$ \\
Kitchen-mixed & $0.01 \pm 0.02$ & $3.66 \pm 0.42$ & $0.02 \pm 0.05$ & $3.87 \pm 0.15$ \\
Top burner & $0.00 \pm 0.00$ & $0.98 \pm 0.02$ & $0.00 \pm 0.00$ & $0.98 \pm 0.02$ \\
\bottomrule
\end{tabular}

\end{table}

\begin{figure}[htbp]
    \centering
    \resizebox{\linewidth}{!}{
        \input{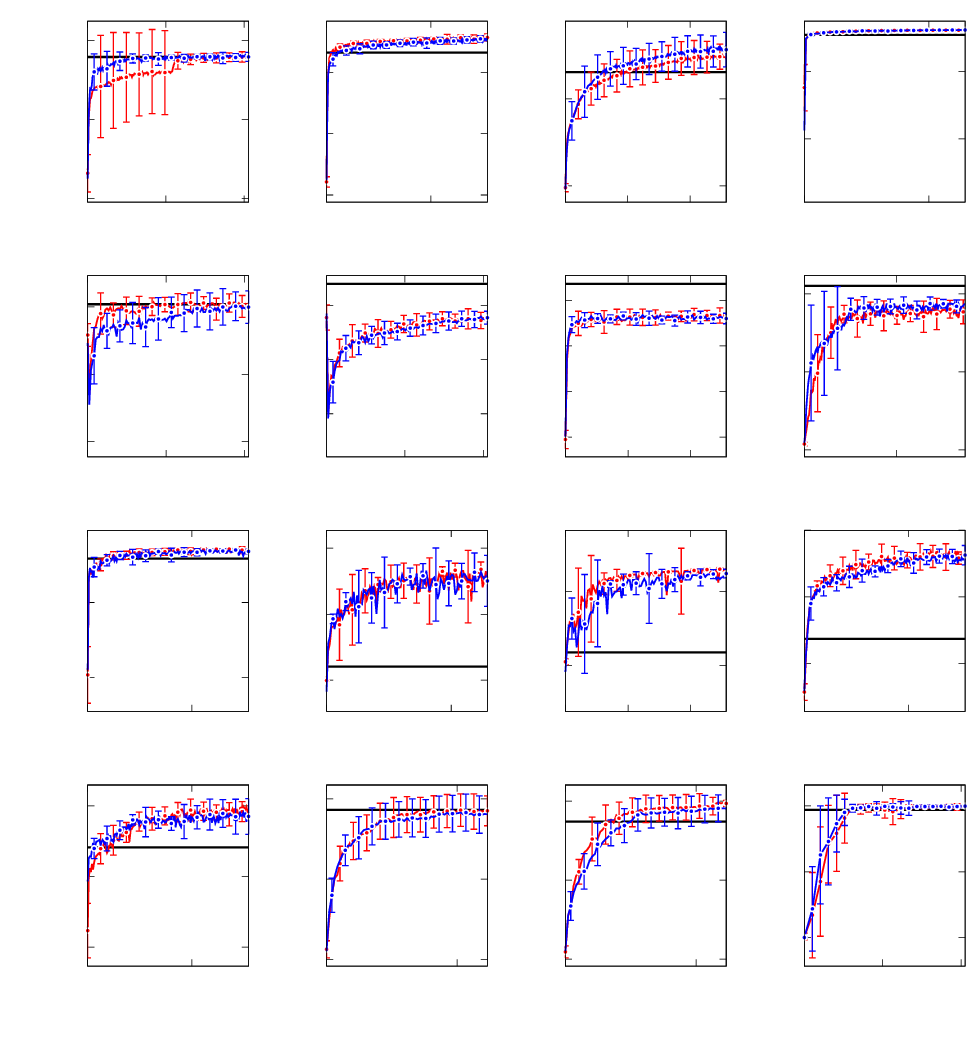}
    }
    \caption{Test performance as a function of training epoch. Blue (Red) indicates without (with) switching penalty while finding motion primitives. Trials without a switching penalty occasionally converge faster across some tasks but tend to reach the same final performance by the end of the budget.}
    \label{fig:training-ablation}
\end{figure}

\begin{figure}[htbp]
    \centering
    \resizebox{\linewidth}{!}{%
        \input{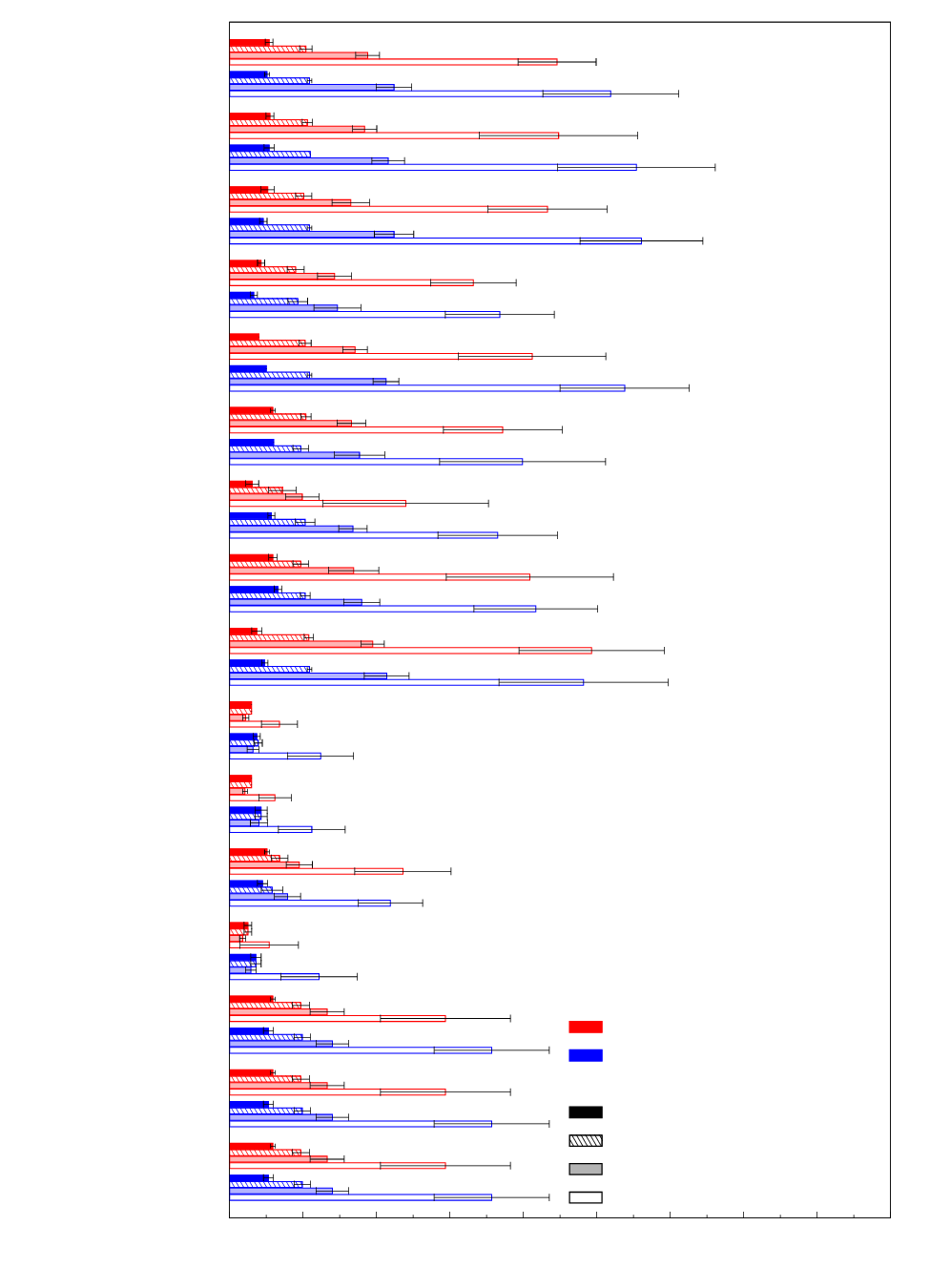}
    }
    \caption{Counts of primitive present in each policy as averaged over 10 trials. Blue (Red) indicates without (with) switching penalty while finding motion primitives. Trials without the switching penalty tend to be more complex across all primitives.}
    \label{fig:complexity-ablation}
\end{figure}

\lstdefinelanguage{policy}{sensitive=true, alsoletter={-:}, alsodigit={.},
  morekeywords=[1]{:accelerate-left,:accelerate-right,:coast,:do-nothing,:fire-left-engine,:fire-main-engine,
    :fire-right-engine,:motion-1,:motion-2,:motion-3,:motion-4,:motion-5,:motion-6,:negative-torque,:no-torque,
    :positive-torque,:push-left,:push-right},
  morekeywords=[2]{define,defpolicy,if,elif,else},
  morekeywords=[3]{acrobot,adroit-door,adroit-hammer,adroit-pen,adroit-relocate,ant,bipedal-walker,cartpole,
    half-cheetah,hopper,humanoid,kitchen,kitchen-mixed,kitchen-top-burner,lunar-lander,mountain-car,pusher,reacher,
    swimmer,walker2d}}
\lstdefinestyle{decisionlist}{language=policy, basicstyle=\ttfamily\scriptsize, columns=fullflexible,
  keepspaces=true, breaklines=true, breakatwhitespace=true, postbreak=\mbox{\hspace{1em}},
  showstringspaces=false, frame=lines, framesep=3pt, xleftmargin=0pt, aboveskip=2pt, belowskip=2pt,
  keywordstyle=[1]\color[RGB]{72,61,139}, keywordstyle=[2]\color[RGB]{147,32,146},
  keywordstyle=[3]\color[RGB]{0,0,255}}

\noindent
\begin{minipage}[t]{\linewidth}
\begin{lstlisting}[style=decisionlist]
(define
  (pred1 (> (- foot-angular-velocity (* (abs x-velocity) thigh-angle)) 3.4016))
  (pred2 (> (abs x-velocity) 2.4834))
  (pred3 (> torso-angle 0.0839))
  (pred4 (< leg-angle -0.3732))
  (pred5 (> (+ (- thigh-angle (* 0.2307 torso-angular-velocity)) (* 0.5382 x-velocity)) 0.7399))
  (pred6 (< torso-height 1.5492))
  (pred7 (> torso-height 1.3137))
  (pred8 (> (+ (+ torso-angle (* 0.0325 (abs x-velocity))) (* 0.3925 leg-angle)) -0.0226))
  (pred9 (> z-velocity -2.1900))
  (pred10 (< foot-angle -0.5612))
  (pred11 (< leg-angular-velocity -2.0288))
  (pred12 (< (+ leg-angle (* 0.1158 x-velocity)) 0.1227))
  (pred13 (> z-velocity 0.6911))
  (pred14 (< (+ leg-angular-velocity (- (angle leg-angular-velocity foot-angle) torso-angle)) -4.9620))
  (pred15 (< (+ leg-angle (* 0.1190 x-velocity)) 0.1469))
  (pred16 (< thigh-angular-velocity -3.5409))
  (pred17 (< (abs thigh-angle) 0.7317))
  (pred18 (< (- torso-angle (* 0.0092 foot-angular-velocity)) -0.1565))
  (pred19 (> torso-angle 0.0640)))

(defpolicy hopper
  (if (and pred1 pred2) :motion-3)
  (elif pred3 :motion-4)
  (elif (and pred4 pred5 pred6) :motion-2)
  (elif (and pred7 pred8 pred9) :motion-4)
  (elif (and pred10 pred11 pred12) :motion-1)
  (elif (and pred13 pred14) :motion-5)
  (elif pred15 :motion-3)
  (elif (and pred16 pred17) :motion-5)
  (elif pred18 :motion-2)
  (elif pred19 :motion-4)
  (else :motion-1))
\end{lstlisting}
{\footnotesize The policy for Hopper with a held-out mean across 100 episodes of $3{,}668$.\par}
\end{minipage}

\bigskip
\noindent
\begin{minipage}[t]{\linewidth}
\begin{lstlisting}[style=decisionlist]
(define
  (pred1 (< (- (- right-foot-angle (* 0.4977 torso-height)) (* 0.1574 left-foot-angle)) -0.0358))
  (pred2 (< right-foot-angle 0.8067))
  (pred3 (< x-velocity -0.0470))
  (pred4 (> left-thigh-angle -0.0712))
  (pred5 (> (- (abs 0.4795) (sqrt (abs left-foot-angular-velocity))) -1.6550))
  (pred6 (< left-foot-angle 0.5650))
  (pred7 (< x-velocity 0.9042))
  (pred8 (> (abs torso-height) 1.1638))
  (pred9 (> z-velocity 0.7358))
  (pred10 (< right-foot-angle 0.8262))
  (pred11 (< (- right-thigh-angular-velocity (* x-velocity (+ right-thigh-angle z-velocity))) -1.2652))
  (pred12 (> left-foot-angle 0.0873))
  (pred13 (< x-velocity 3.0861))
  (pred14 (< left-foot-angle 0.7474))
  (pred15 (> right-foot-angle 0.8269))
  (pred16 (> z-velocity 0.1809))
  (pred17 (> (p-div left-foot-angular-velocity -1.1606) 0.2294))
  (pred18 (< (- right-thigh-angular-velocity (* 134.1759 (+ right-thigh-angle z-velocity))) 15.7290))
  (pred19 (> left-thigh-angle -0.0926))
  (pred20 (< right-thigh-angle 0.0509)))

(defpolicy walker2d
  (if (and pred1 pred2) :motion-4)
  (elif pred3 :motion-5)
  (elif (and pred4 pred5 pred6) :motion-5)
  (elif (and pred7 pred8) :motion-2)
  (elif (and pred9 pred10 pred11) :motion-3)
  (elif (and pred12 pred13 pred14) :motion-5)
  (elif (and pred15 pred16) :motion-2)
  (elif (and pred17 pred18) :motion-5)
  (elif (and pred19 pred20) :motion-1)
  (else :motion-2))
\end{lstlisting}
{\footnotesize The policy for Walker2d with a held-out mean across 100 episodes of $5{,}302$.\par}
\end{minipage}

\bigskip
\noindent
\begin{minipage}[t]{\linewidth}
\begin{lstlisting}[style=decisionlist]
(define
  (pred1 (> (+ torso-angle back-foot-angle) 0.7093))
  (pred2 (> back-foot-angular-velocity 1.1436))
  (pred3 (> (+ front-shin-angle front-foot-angular-velocity) 19.4990))
  (pred4 (> (abs back-shin-angle) 0.4273))
  (pred5 (< front-foot-angular-velocity -3.1508))
  (pred6 (> (- (- torso-angle (* 6.7217 torso-height)) (* 0.0094 back-thigh-angle)) 0.0626))
  (pred7 (< front-foot-angle -0.3155))
  (pred8 (> (- (- torso-angle (* 6.8929 torso-height)) (* 0.0047 (- back-foot-angular-velocity))) 0.1694))
  (pred9 (< (+ front-shin-angular-velocity (- (- front-foot-angle (* 7.2016 torso-height)) (* 0.0218 (- back-foot-angular-velocity)))) 4.4780))
  (pred10 (< back-thigh-angle -0.8664))
  (pred11 (> front-shin-angle -0.2095))
  (pred12 (> (- (- (- front-foot-angle torso-height) (* 13.2004 torso-height)) (* 0.0100 (- -1.5074))) 0.0162))
  (pred13 (< back-thigh-angle -0.7097))
  (pred14 (< back-foot-angular-velocity 13.6064))
  (pred15 (> (- (- (- z-velocity torso-height) (* 4.6909 torso-height)) (* 0.0163 (sqrt (abs back-shin-angle)))) 0.4063))
  (pred16 (> (+ torso-height (* 0.2739 back-thigh-angle)) 0.0905))
  (pred17 (< (+ front-shin-angular-velocity (* 0.4352 front-thigh-angle)) 4.1479)))

(defpolicy half-cheetah
  (if (and pred1 pred2) :motion-5)
  (elif (and pred3 pred4) :motion-3)
  (elif pred5 :motion-5)
  (elif (and pred6 pred7) :motion-1)
  (elif (and pred8 pred9) :motion-4)
  (elif (and pred10 pred11) :motion-1)
  (elif pred12 :motion-4)
  (elif (and pred13 pred14) :motion-1)
  (elif pred15 :motion-4)
  (elif (and pred16 pred17) :motion-5)
  (else :motion-3))
\end{lstlisting}
{\footnotesize The policy for HalfCheetah with a held-out mean across 100 episodes of $9{,}378$.\par}
\end{minipage}

\bigskip
\noindent
\begin{minipage}[t]{\linewidth}
\begin{lstlisting}[style=decisionlist]
(define
  (pred1 (< (+ (+ joint2-angle joint2-angular-velocity) (* 0.3321 tip-angular-velocity)) -3.4861))
  (pred2 (> tip-angular-velocity 0.0452))
  (pred3 (< (+ tip-angular-velocity joint2-angular-velocity) -0.6501))
  (pred4 (> tip-angular-velocity -0.2852))
  (pred5 (< (abs tip-angular-velocity) 2.2277))
  (pred6 (> tip-angular-velocity -0.0499))
  (pred7 (between (angle (* 0.7656 tip-angular-velocity) joint2-angle) -5.8381 104.4973))
  (pred8 (between (angle (* 1.3156 y-velocity) joint1-angular-velocity) 99.8867 -115.0106))
  (pred9 (> (+ joint2-angle joint2-angular-velocity) 3.0530))
  (pred10 (between (angle (* 1.6820 y-velocity) joint1-angular-velocity) 100.8340 -124.2093))
  (pred11 (< (angle tip-angle (angle joint1-angular-velocity tip-angle)) -12.0551))
  (pred12 (> (* 0.5416 tip-angular-velocity) 0.0735)))

(defpolicy swimmer
  (if (and pred1 pred2) :motion-1)
  (elif (and pred3 pred4 pred5) :motion-1)
  (elif (and pred6 pred7 pred8) :motion-2)
  (elif pred9 :motion-4)
  (elif (and pred10 pred11) :motion-2)
  (elif pred12 :motion-5)
  (else :motion-3))
\end{lstlisting}
{\footnotesize The policy for Swimmer with a held-out mean across 100 episodes of $362.6$.\par}
\end{minipage}

\bigskip
\noindent
\begin{minipage}[t]{\linewidth}
\begin{lstlisting}[style=decisionlist]
(define
  (pred1 (> target-y 0.0340))
  (pred2 (> fingertip-minus-target-x 0.0547))
  (pred3 (< (angle (expt sin-joint1 2) fingertip-minus-target-x) 66.8601))
  (pred4 (> (abs (angle cos-joint1 target-y)) 88.4500))
  (pred5 (> cos-joint1 0.7897))
  (pred6 (> (norm target-x target-y) 0.0316))
  (pred7 (< (- cos-joint0) -0.6274))
  (pred8 (> (- joint0-angular-velocity (* 10.2587 sin-joint0)) -2.6069))
  (pred9 (< sin-joint0 -0.4311))
  (pred10 (< joint0-angular-velocity 2.5828))
  (pred11 (> (+ (+ joint1-angular-velocity (* 7.5741 cos-joint0)) (* 25.5120 fingertip-minus-target-x)) -11.1597))
  (pred12 (> (abs (angle (p-div sin-joint1 sin-joint1) target-y)) 89.2170))
  (pred13 (< sin-joint0 0.4648))
  (pred14 (> fingertip-minus-target-y 0.0610))
  (pred15 (< (angle (- target-y) cos-joint0) -47.6930))
  (pred16 (> cos-joint1 0.8749)))

(defpolicy reacher
  (if (and pred1 pred2 pred3) :motion-4)
  (elif (and pred4 pred5 pred6) :motion-3)
  (elif (and pred7 pred8) :motion-2)
  (elif pred9 :motion-2)
  (elif (and pred10 pred11) :motion-1)
  (elif pred12 :motion-2)
  (elif pred13 :motion-3)
  (elif (and pred14 pred15) :motion-4)
  (elif pred16 :motion-4)
  (else :motion-1))
\end{lstlisting}
{\footnotesize The policy for Reacher with a held-out mean across 100 episodes of $-4.77$.\par}
\end{minipage}

\bigskip
\noindent
\begin{minipage}[t]{\linewidth}
\begin{lstlisting}[style=decisionlist]
(define
  (pred1 (< (p-div (p-div -0.4743 0.4372) shoulder-lift-angle) -27589.6874))
  (pred2 (< object-y -0.2170))
  (pred3 (> (norm object-x object-y) 0.4207))
  (pred4 (> (norm fingertip-x fingertip-y fingertip-z) 0.9728))
  (pred5 (< elbow-flex-velocity -0.5509))
  (pred6 (< (+ forearm-roll-velocity (* 2.0129 wrist-roll-angle)) 1.2820))
  (pred7 (< (p-div (p-div -0.4951 0.4356) shoulder-lift-angle) -4.5968))
  (pred8 (> (- (p-div shoulder-lift-velocity 2.4268) (- shoulder-lift-angle -0.0941)) -1.0110))
  (pred9 (> (abs object-x) 0.5176))
  (pred10 (< (angle (* 0.4676 fingertip-x) (angle object-y wrist-roll-velocity)) 179.7953))
  (pred11 (> elbow-flex-velocity -0.0309))
  (pred12 (< (- (+ shoulder-pan-angle (* elbow-flex-velocity elbow-flex-velocity)) (* 7.1773 fingertip-z)) 1.8975))
  (pred13 (between (angle (* (angle 0.9517 wrist-flex-velocity) wrist-flex-velocity) shoulder-lift-velocity) 157.0843 -126.5096))
  (pred14 (> (norm object-x object-y) 0.4325))
  (pred15 (> shoulder-pan-velocity -0.0030))
  (pred16 (< (+ upper-arm-roll-velocity (* 1.1175 wrist-flex-angle)) -0.1992)))

(defpolicy pusher
  (if (and pred1 pred2 pred3) :motion-6)
  (elif pred4 :motion-1)
  (elif (and pred5 pred6) :motion-3)
  (elif pred7 :motion-4)
  (elif pred8 :motion-2)
  (elif pred9 :motion-5)
  (elif (and pred10 pred11 pred12) :motion-5)
  (elif (and pred13 pred12) :motion-4)
  (elif (and pred14 pred15) :motion-4)
  (elif pred16 :motion-4)
  (else :motion-2))
\end{lstlisting}
{\footnotesize The policy for Pusher with a held-out mean across 100 episodes of $-32.4$.\par}
\end{minipage}

\bigskip
\noindent
\begin{minipage}[t]{\linewidth}
\begin{lstlisting}[style=decisionlist]
(define
  (pred1 (> torso-height 0.6199))
  (pred2 (< front-left-hip-velocity -1.0334))
  (pred3 (> back-left-hip-velocity 5.7813))
  (pred4 (< (- (+ back-left-ankle-velocity back-right-ankle-velocity) front-left-ankle-velocity) 0.6289))
  (pred5 (> back-left-hip-angle -0.3718))
  (pred6 (> front-left-hip-angle -0.4705))
  (pred7 (< (- (- front-right-hip-angle (* 0.2505 z-velocity)) (* 11.0977 torso-qy)) -0.7163))
  (pred8 (> (p-div back-left-hip-angle -0.2316) -0.9384))
  (pred9 (< (- roll-rate (* 0.7433 back-left-hip-angle)) 0.4588)))

(defpolicy ant
  (if pred1 :motion-1)
  (elif pred2 :motion-1)
  (elif pred3 :motion-1)
  (elif (and pred4 pred5 pred6) :motion-2)
  (elif pred7 :motion-1)
  (elif (and pred8 pred9) :motion-3)
  (elif pred8 :motion-1)
  (else :motion-4))
\end{lstlisting}
{\footnotesize The policy for Ant with a held-out mean across 100 episodes of $5{,}720$.\par}
\end{minipage}

\bigskip
\noindent
\begin{minipage}[t]{\linewidth}
\begin{lstlisting}[style=decisionlist]
(define
  (pred1 (> right-knee-angle -0.0897))
  (pred2 (> (norm torso-qx torso-qy torso-qz) 0.5622))
  (pred3 (> right-knee-angle 0.0135))
  (pred4 (> (+ (- left-elbow-angle (* 1.1655 left-shin-inertia-2)) (* 1.2888 right-thigh-inertia-8)) -0.4922))
  (pred5 (between (angle right-foot-inertia-8 left-foot-contact-force-0) -150.5126 -53.5966))
  (pred6 (< (- (- right-knee-angle (* 0.1132 (* right-shin-inertia-3 torso-height))) (* 0.0112 right-elbow-angle)) -1.0491))
  (pred7 (> right-thigh-comvel-4 0.2009))
  (pred8 (between (angle (* lwaist-inertia-8 left-hip-z-angle) (abs torso-inertia-2)) -153.7896 20.7418))
  (pred9 (> (- right-thigh-inertia-1 (* 6.9913 left-thigh-inertia-3)) 0.4672))
  (pred10 (< (- (- right-knee-angle (* 0.1144 pelvis-comvel-2)) (* 0.0107 left-knee-velocity)) -1.0005))
  (pred11 (< (- (* 0.0100 right-thigh-comvel-4) (p-div left-foot-contact-force-3 right-lower-arm-inertia-8)) -10.3673))
  (pred12 (> (+ (* right-hip-y-velocity -0.6912) right-lower-arm-inertia-2) 2.3075))
  (pred13 (< left-knee-velocity 18.6165)))

(defpolicy humanoid
  (if (and pred1 pred2) :motion-3)
  (elif (and pred3 (not pred4)) :motion-3)
  (elif (and (not pred5) pred6 pred7) :motion-1)
  (elif pred8 :motion-2)
  (elif pred9 :motion-5)
  (elif (and pred10 pred11) :motion-1)
  (elif pred12 :motion-2)
  (elif pred13 :motion-4)
  (else :motion-2))
\end{lstlisting}
{\footnotesize The policy for Humanoid with a held-out mean across 100 episodes of $9{,}976$.\par}
\end{minipage}

\bigskip
\noindent
\begin{minipage}[t]{\linewidth}
\begin{lstlisting}[style=decisionlist]
(define
  (pred1 (between (angle (* 2.7376 cart-velocity) (+ cart-position pole-angular-velocity)) -62.0642 136.4117))
  (pred2 (< (+ pole-angle (* 0.1165 pole-angular-velocity)) 0.0017))
  (pred3 (> (+ pole-angular-velocity pole-angle) -0.0854))
  (pred4 (between (angle (* 7.2509 pole-angle) pole-angular-velocity) 165.2007 -74.3086))
  (pred5 (between (angle (* 6.4616 (* (- pole-angle cart-velocity) pole-angular-velocity)) (+ pole-angular-velocity pole-angle)) -148.9216 -94.4409)))

(defpolicy cartpole
  (if (and (not pred1) pred2) :push-left)
  (elif (and pred3 (not pred4) (not pred5)) :push-right)
  (else :push-left))
\end{lstlisting}
{\footnotesize The policy for CartPole with a held-out mean across 100 episodes of $500.0$.\par}
\end{minipage}

\bigskip
\noindent
\begin{minipage}[t]{\linewidth}
\begin{lstlisting}[style=decisionlist]
(define
  (pred1 (< (+ (+ car-position car-velocity) (+ car-position car-velocity)) -0.3993))
  (pred2 (> (- (+ car-position (* car-position car-velocity)) (* 51.1200 car-velocity)) -0.4679))
  (pred3 (< (- car-velocity (* 0.1906 (+ (+ car-position car-velocity) car-velocity))) 0.1709))
  (pred4 (< (+ car-velocity (* 0.0675 car-position)) -0.0777))
  (pred5 (< car-velocity 0.0045))
  (pred6 (< (+ (+ car-position car-velocity) (+ car-position car-velocity)) -0.6569))
  (pred7 (< (- car-velocity (* 0.1703 (+ (+ car-position car-position) car-velocity))) 0.1540)))

(defpolicy mountain-car
  (if (and pred1 pred2 pred3) :accelerate-left)
  (elif (and pred4 pred5) :coast)
  (elif (and pred6 pred7) :accelerate-left)
  (else :accelerate-right))
\end{lstlisting}
{\footnotesize The policy for MountainCar with a held-out mean across 100 episodes of $-98.4$.\par}
\end{minipage}

\bigskip
\noindent
\begin{minipage}[t]{\linewidth}
\begin{lstlisting}[style=decisionlist]
(define
  (pred1 (< sin-theta1 -0.0337))
  (pred2 (< (- cos-theta1 (* 0.2859 theta1-angular-velocity)) 0.9351))
  (pred3 (< cos-theta2 -0.6765))
  (pred4 (< sin-theta1 0.8774))
  (pred5 (< (+ cos-theta2 (* 0.2439 sin-theta1)) -1.1173))
  (pred6 (< theta2-angular-velocity 0.1961))
  (pred7 (> theta1-angular-velocity -0.4920))
  (pred8 (< (- cos-theta1 (* 0.3111 theta1-angular-velocity)) 0.3092))
  (pred9 (between (angle (* 18.5282 cos-theta1) theta2-angular-velocity) 15.2184 144.4511))
  (pred10 (< theta2-angular-velocity 0.3341))
  (pred11 (< sin-theta2 0.9981))
  (pred12 (< (- cos-theta2 (* 1.6043 sin-theta2)) 1.0051)))

(defpolicy acrobot
  (if (and (not pred1) (not pred2)) :positive-torque)
  (elif (and pred3 (not pred4)) :positive-torque)
  (elif pred5 :negative-torque)
  (elif (and pred6 pred7) :negative-torque)
  (elif (and pred8 pred9) :no-torque)
  (elif (and pred10 pred11 pred12) :negative-torque)
  (else :positive-torque))
\end{lstlisting}
{\footnotesize The policy for Acrobot with a held-out mean across 100 episodes of $-72.6$.\par}
\end{minipage}

\bigskip
\noindent
\begin{minipage}[t]{\linewidth}
\begin{lstlisting}[style=decisionlist]
(define
  (pred1 (< lander-angle -0.1032))
  (pred2 (> (- (- x-velocity (* 0.7291 lander-angle)) (* 2.3967 angular-velocity)) 0.1732))
  (pred3 (between (angle right-leg-contact (+ lander-angle angular-velocity)) -16.3818 76.6411))
  (pred4 (> (- angular-velocity (* 0.6826 x-velocity)) -0.0269))
  (pred5 (< (- (- x-velocity angular-velocity) angular-velocity) 0.4687))
  (pred6 (< (+ y (* 4.2702 y-velocity)) -0.7985))
  (pred7 (> (+ angular-velocity (* 3.4627 lander-angle)) -0.3170))
  (pred8 (< (- right-leg-contact (* (* 5.3354 y-velocity) x-velocity)) 0.8131))
  (pred9 (< (+ y (* 4.2624 y-velocity)) -0.8781))
  (pred10 (< y-velocity -0.0252))
  (pred11 (> (- (- x-velocity (* 0.9284 lander-angle)) (* 2.2148 angular-velocity)) 0.0975))
  (pred12 (> (+ (- right-leg-contact (* 4.9244 lander-angle)) (* 5.9569 x)) -0.1785))
  (pred13 (< (- right-leg-contact (* 2.2458 x-velocity)) 0.4768)))

(defpolicy lunar-lander
  (if (and pred1 pred2) :fire-left-engine)
  (elif (and pred3 pred4) :fire-right-engine)
  (elif (and pred5 pred6 pred7) :fire-main-engine)
  (elif (and pred8 pred9) :fire-main-engine)
  (elif (and pred10 pred11) :fire-left-engine)
  (elif (not pred12) :fire-right-engine)
  (elif pred13 :fire-left-engine)
  (else :do-nothing))
\end{lstlisting}
{\footnotesize The policy for LunarLander with a held-out mean across 100 episodes of $263.6$.\par}
\end{minipage}

\bigskip
\noindent
\begin{minipage}[t]{\linewidth}
\begin{lstlisting}[style=decisionlist]
(define
  (pred1 (< (angle (- hull-angle lidar-8) (- knee1-angle lidar-7)) -142.9046))
  (pred2 (< (- lidar-7 hip2-angle) 0.8607))
  (pred3 (> (+ hip2-speed (* 0.6567 lidar-6)) 0.8982))
  (pred4 (< (abs x-velocity) 0.1998))
  (pred5 (< knee1-angle 0.0204))
  (pred6 (> (- knee2-speed (* 3.4578 hip2-speed)) -2.6599))
  (pred7 (> x-velocity 0.0300))
  (pred8 (> (+ hip1-speed (* 4.8084 hip2-angle)) 0.3148))
  (pred9 (> (+ knee2-speed (* 3.1111 leg2-contact)) -0.3957))
  (pred10 (< (- lidar-6 knee2-angle) 0.1160))
  (pred11 (> (- (p-div lidar-3 lidar-2) knee1-angle) 1.5981))
  (pred12 (< (p-div 0.4034 knee2-speed) 0.4976))
  (pred13 (> hip1-speed -0.4565))
  (pred14 (> knee2-angle -0.2207))
  (pred15 (< (- x-velocity (* 26.7785 hull-angular-velocity)) 0.0622))
  (pred16 (< (+ (- hip1-angle (* 0.7177 hull-angle)) (* 3.8558 lidar-1)) 2.1242))
  (pred17 (> hip1-speed -0.2421))
  (pred18 (between (angle (* 3.3116 x-velocity) leg2-contact) 66.4538 128.9908)))

(defpolicy bipedal-walker
  (if (and pred1 pred2 pred3) :motion-3)
  (elif (and pred4 pred5 pred6) :motion-1)
  (elif (and pred7 pred8 pred9) :motion-3)
  (elif pred10 :motion-2)
  (elif (and pred11 pred12 pred13) :motion-2)
  (elif (and pred14 pred12) :motion-4)
  (elif (and pred15 pred16) :motion-3)
  (elif (and pred17 (not pred18)) :motion-3)
  (else :motion-4))
\end{lstlisting}
{\footnotesize The policy for BipedalWalker with a held-out mean across 100 episodes of $344.7$.\par}
\end{minipage}

\bigskip
\noindent
\begin{minipage}[t]{\linewidth}
\begin{lstlisting}[style=decisionlist]
(define
  (pred1 (< (angle (p-div (- lfj1 palm-y) 0.5885) (- handle-x handle-y)) 80.1094))
  (pred2 (< thj0 -0.4344)))

(defpolicy adroit-door
  (if pred1 :motion-1)
  (elif pred2 :motion-2)
  (else :motion-3))
\end{lstlisting}
{\footnotesize The policy for Adroit Door with a held-out mean across 100 episodes of $3{,}356$.\par}
\end{minipage}

\bigskip
\noindent
\begin{minipage}[t]{\linewidth}
\begin{lstlisting}[style=decisionlist]
(define
  (pred1 (> (+ hammer-z (* 3.0166 nail-head-x)) 0.2518))
  (pred2 (< (sqrt (abs (* (* ffj0 mfj2) lfj0))) 0.6493)))

(defpolicy adroit-hammer
  (if pred1 :motion-2)
  (elif pred2 :motion-1)
  (else :motion-3))
\end{lstlisting}
{\footnotesize The policy for Adroit Hammer with a held-out mean across 100 episodes of $18{,}701$.\par}
\end{minipage}

\bigskip
\noindent
\begin{minipage}[t]{\linewidth}
\begin{lstlisting}[style=decisionlist]
(define
  (pred1 (< (+ lfj4 (* 0.1620 pen-wx)) -0.4122))
  (pred2 (> (+ thj4 (+ mfj0 pen-dir-y)) 1.4221))
  (pred3 (< pen-minus-target-x 0.0317))
  (pred4 (< pen-dir-z 0.0087))
  (pred5 (< lfj0 0.9034))
  (pred6 (< (+ wrj0 (* 0.7290 ffj1)) -0.4420))
  (pred7 (< ffj1 0.9518)))

(defpolicy adroit-pen
  (if pred1 :motion-5)
  (elif (and pred2 pred3) :motion-2)
  (elif (and pred4 pred5) :motion-3)
  (elif pred6 :motion-5)
  (elif pred7 :motion-1)
  (else :motion-4))
\end{lstlisting}
{\footnotesize The policy for Adroit Pen with a held-out mean across 100 episodes of $4{,}672$.\par}
\end{minipage}

\bigskip
\noindent
\begin{minipage}[t]{\linewidth}
\begin{lstlisting}[style=decisionlist]
(define
  (pred1 (< (- mfj0 lfj3) 0.7589))
  (pred2 (> (- ffj0 (expt thj1 2)) 0.8567)))

(defpolicy adroit-relocate
  (if pred1 :motion-2)
  (elif pred2 :motion-4)
  (else :motion-3))
\end{lstlisting}
{\footnotesize The policy for Adroit Relocate with a held-out mean across 100 episodes of $5{,}049$.\par}
\end{minipage}

\bigskip
\noindent
\begin{minipage}[t]{\linewidth}
\begin{lstlisting}[style=decisionlist]
(define
  (pred1 (< panda-joint6-angle 0.9626))
  (pred2 (< (- (+ knob2-velocity (* 1.1900 finger2-velocity)) (expt -0.2226 2)) -0.2180))
  (pred3 (between (angle (* 1034.4153 left-hinge-cabinet-angle) knob4-velocity) 102.1846 -90.3997))
  (pred4 (between (angle (norm kettle-vx kettle-vy kettle-vz) right-hinge-cabinet-angle) -121.8511 44.6983))
  (pred5 (> panda-joint7-angle -0.5280))
  (pred6 (< panda-joint1-angle -0.9794))
  (pred7 (> kettle-qx 0.0376))
  (pred8 (> (+ panda-joint1-velocity (* panda-joint1-velocity light-switch-angle)) 0.1494))
  (pred9 (< light-switch-velocity -0.0392))
  (pred10 (> (- (+ panda-joint6-velocity panda-joint1-velocity) (- kettle-vz)) -0.0933))
  (pred11 (< (- finger1-position (* 0.0274 panda-joint1-angle)) 0.0706))
  (pred12 (< light-switch-velocity 0.0147))
  (pred13 (< light-joint-velocity 0.0005837)))

(defpolicy kitchen
  (if pred1 :motion-6)
  (elif pred2 :motion-2)
  (elif pred3 :motion-2)
  (elif (and pred4 pred5 pred6) :motion-5)
  (elif pred7 :motion-1)
  (elif (and pred8 (not pred9) pred10) :motion-3)
  (elif (and pred11 (not pred12)) :motion-6)
  (elif (not pred12) :motion-4)
  (elif (not pred13) :motion-6)
  (else :motion-4))
\end{lstlisting}
{\footnotesize The policy for Kitchen with a held-out mean across 100 episodes of $3.98$.\par}
\end{minipage}

\bigskip
\noindent
\begin{minipage}[t]{\linewidth}
\begin{lstlisting}[style=decisionlist]
(define
  (pred1 (< panda-joint6-angle 0.8250))
  (pred2 (< (+ (- finger1-position (* (* 0.7655 right-hinge-cabinet-angle) panda-joint7-angle)) panda-joint7-velocity) -0.0189))
  (pred3 (< (- panda-joint2-velocity (* 0.5531 light-switch-velocity)) 0.0867))
  (pred4 (> right-hinge-cabinet-velocity 0.0085))
  (pred5 (> (+ (sqrt (abs panda-joint6-angle)) (sqrt (abs 1.6391))) 2.7113))
  (pred6 (> panda-joint3-velocity -0.0587))
  (pred7 (< (- microwave-velocity (* (+ knob4-velocity (* 0.8242 panda-joint6-velocity)) right-hinge-cabinet-angle)) -0.0043))
  (pred8 (< light-switch-angle -0.6869))
  (pred9 (between (angle left-hinge-cabinet-angle light-joint-velocity) 162.5492 -107.7113))
  (pred10 (> kettle-qx 0.0630))
  (pred11 (< (+ light-switch-angle panda-joint1-velocity) 0.1916))
  (pred12 (< panda-joint7-angle 2.7835)))

(defpolicy kitchen-mixed
  (if pred1 :motion-2)
  (elif (and pred2 pred3 pred4) :motion-1)
  (elif (and pred5 pred6) :motion-3)
  (elif pred7 :motion-5)
  (elif pred8 :motion-4)
  (elif pred9 :motion-4)
  (elif pred10 :motion-3)
  (elif pred11 :motion-2)
  (elif pred12 :motion-5)
  (else :motion-6))
\end{lstlisting}
{\footnotesize The policy for Kitchen-mixed with a held-out mean across 100 episodes of $3.98$.\par}
\end{minipage}

\bigskip
\noindent
\begin{minipage}[t]{\linewidth}
\begin{lstlisting}[style=decisionlist]
(define
  (pred1 (< (+ left-hinge-cabinet-angle (* 8.281e-05 (- panda-joint6-angle -0.3936))) -0.0064))
  (pred2 (< (- (* 0.4470 right-hinge-cabinet-angle) (* 9.5037 right-hinge-cabinet-velocity)) -0.6946))
  (pred3 (< (+ left-hinge-cabinet-velocity (* 0.0012 finger1-velocity)) 5.085e-05))
  (pred4 (> (angle (abs 0.6873) light-joint-velocity) 90.0997))
  (pred5 (< (+ light-switch-angle (* 11.7204 kettle-wz)) -0.0060))
  (pred6 (< (+ light-switch-angle (* knob4-velocity kettle-wz)) -0.7037))
  (pred7 (> kettle-qx 0.0371))
  (pred8 (> (- (expt kettle-vy 2) (+ right-hinge-cabinet-angle panda-joint2-velocity)) 0.2204))
  (pred9 (< knob4-velocity -0.0800))
  (pred10 (< (+ left-hinge-cabinet-angle (* 7.253e-05 panda-joint3-angle)) -0.0064))
  (pred11 (< (- right-hinge-cabinet-angle (* 1.3320 panda-joint6-velocity)) 1.5818))
  (pred12 (> (- (- microwave-velocity) (* 0.2246 panda-joint4-velocity)) 0.0266))
  (pred13 (> (+ top-left-burner-position (* 0.0828 light-joint-position)) -0.0041))
  (pred14 (> (+ panda-joint4-angle (* 197.9562 bottom-left-burner-position)) -2.9103))
  (pred15 (> (* 0.2331 panda-joint4-velocity) 0.0252))
  (pred16 (> (sqrt (abs (+ kettle-qy panda-joint7-angle))) 1.3322)))

(defpolicy kitchen-top-burner
  (if pred1 :motion-4)
  (elif (and pred2 pred3) :motion-5)
  (elif (and pred4 pred5) :motion-2)
  (elif pred6 :motion-2)
  (elif pred7 :motion-3)
  (elif pred8 :motion-1)
  (elif pred9 :motion-6)
  (elif (and pred10 pred11) :motion-4)
  (elif (and pred12 pred13 pred14) :motion-5)
  (elif (and pred15 pred16) :motion-3)
  (else :motion-1))
\end{lstlisting}
{\footnotesize The policy for Top burner with a held-out mean across 100 episodes of $1.00$.\par}
\end{minipage}

\bigskip

\end{document}